\documentclass{article}
\usepackage{iclr2024_conference_mod,times}
\preprintcopy
\usepackage{amsmath,amssymb,amsthm}
\usepackage[T1]{fontenc}
\usepackage[utf8]{inputenc}
\usepackage[hidelinks]{hyperref}
\usepackage{url,xurl}
\usepackage{graphicx,booktabs,float,multirow,tabularx,xltabular}
\newcolumntype{Y}{>{\raggedright\arraybackslash}X}
\usepackage{inconsolata}
\usepackage{enumitem}
\usepackage{microtype}
\usepackage[capitalize]{cleveref}
\usepackage[compact]{titlesec}
\usepackage{caption,subcaption}
\usepackage{xcolor}
\newtheorem{proposition}{Proposition}
\newtheorem{corollary}{Corollary}
\newcommand{\ci}[2]{[#1,\,#2]}
\newcommand{\head}[2]{L#1h#2}
\newcommand{\fig}[1]{\IfFileExists{#1}{\includegraphics[width=\linewidth]{#1}}{\fbox{\parbox[c][1.6in][c]{0.96\linewidth}{\centering\path{#1} (pending)}}}}

\title{A Circuit for In-Context Learning}
\author{Andrew Gao\thanks{Equal contribution. Correspondence to gaodrew@stanford.edu. Code, batteries and an interactive walkthrough: \url{https://github.com/andrewgcodes/in-context-learning-circuit}.} \\
Stanford University
\And
William Gao\footnotemark[1] \\
Stanford University}

\begin{document}
\maketitle
\vspace{-7mm}\noindent\begin{minipage}{\linewidth}\centering
\includegraphics[width=0.78\linewidth]{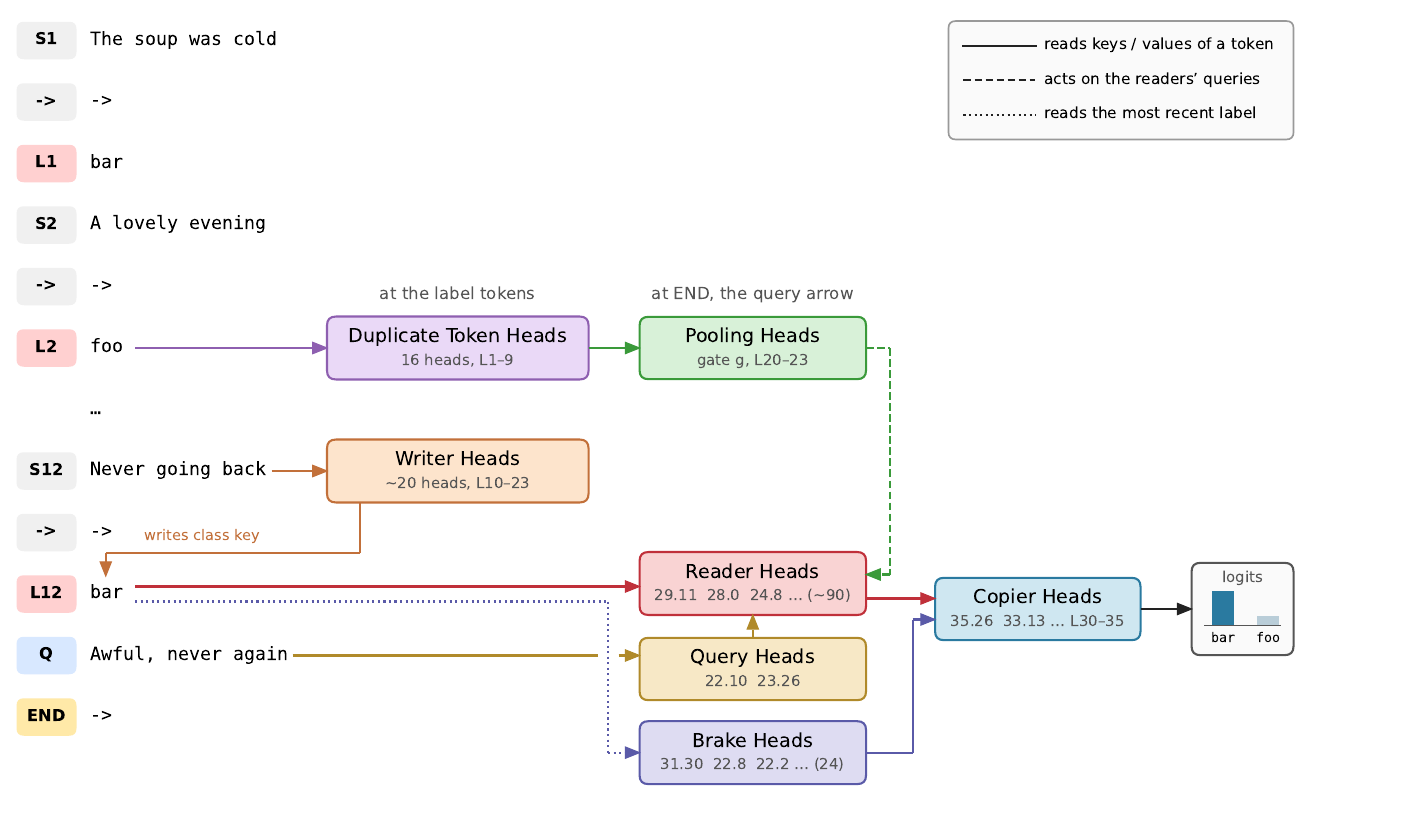}
\end{minipage}\vspace{-3mm}

\begin{abstract}
In-context learning (ICL), a pretrained language model's ability to perform a task specified by demonstrations in its prompt, needs no training objective and no parameter updates. With arbitrary labels the model must acquire a mapping it cannot have memorized, so the learning happens within a single forward pass. Prior work found candidate parts but not how they compose. We give the composition for few-shot sentence classification with arbitrary labels in three 3--4B-parameter transformers. The model implements a key-value memory over label-token positions: each label token's residual stream encodes the label's identity, and middle-layer attention heads write into it the class of the preceding demonstration as its key. At the query, upper-layer heads match the query's class against the stored keys and copy out the matching labels. An early-layer signal encoding whether any label has repeated gates this readout; in its absence the readers attend to the beginning of the prompt and no label is favored, a default that softmax normalization makes unavoidable. A separate set of heads suppresses the most recently seen label, removing a recency bias. Each step is established causally by activation patching between minimal-pair prompts, head ablations against random-head baselines, and steering with the model's own signals. Cutting the connections shows that sixty percent of a class flip passes through the readers and that a few amplified readers cannot replace the rest. A softmax model of the readout with two identifiable constants per head, fit on one prompt distribution, predicts another. The computation is conserved across the three models; which heads fill each role varies. The named heads alone reproduce the behaviour, and a transformer trained from scratch grows the same parts, the gate only when retrieval is penalised. The circuit transfers to semantic concepts; with the concept's own words as labels, the same reader and brake types carry the part of the margin that follows the demonstrations, while the part that follows word meaning lies outside it. The same parts, with a causal gate, appear from 0.6B to 8B in one family; across that range the brakes' hold on word meaning rises with the flipped-label margin, the behaviour \citet{wei2023larger} found only in large models, and the base model's heads still carry retrieval after instruction tuning. ICL in this setting is a gated associative lookup built from a few identifiable component types, traced causally from demonstrations to prediction.
\end{abstract}

\section{Introduction}\label{sec:intro}
Show a base language model a dozen lines like these,
\begin{quote}\small\ttfamily
The soup was cold and the waiter rude -> bar\\
A gorgeous evening with old friends -> foo\\
\ldots\\
I would not go back to that place ->
\end{quote}
and it answers \texttt{bar}. Nothing in pretraining taught it that \texttt{foo} means a pleasant evening or that \texttt{bar} means a bad meal. The model worked that out from the prompt, in a single forward pass, without any change to its weights. This is few-shot in-context learning~\citep{brown2020fewshot}, one of the abilities that turned language models into general-purpose tools. Labelling sentences is its simplest form, and it is the form we study here; instructions, long-form tasks and binding among many classes are outside the scope of this paper.

Several pieces of the mechanism are already known. Induction heads copy the token that followed an earlier occurrence of the current token~\citep{olsson2022induction}. Work on task vectors and function vectors~\citep{hendel2023taskvectors,todd2023fv} has shown that a handful of heads carry a compact description of the demonstrated task to the query. Label tokens are where task information accumulates~\citep{cho2024circuit,zheng2024labelwords,hu2025addition}, and the recency bias of in-context learning has been traced to query--key similarity with label positions~\citep{yu2024qk}. What has been missing is an account of how these pieces fit together into one working algorithm. It was not known exactly what gets written at the label tokens, or on which side of attention it is later read; which heads read it, and what switches that read on and off; how the result reaches the logits; or which of the parts are single heads one can name and which exist only as populations that act in sum.

\begin{figure}[t]\centering
\fig{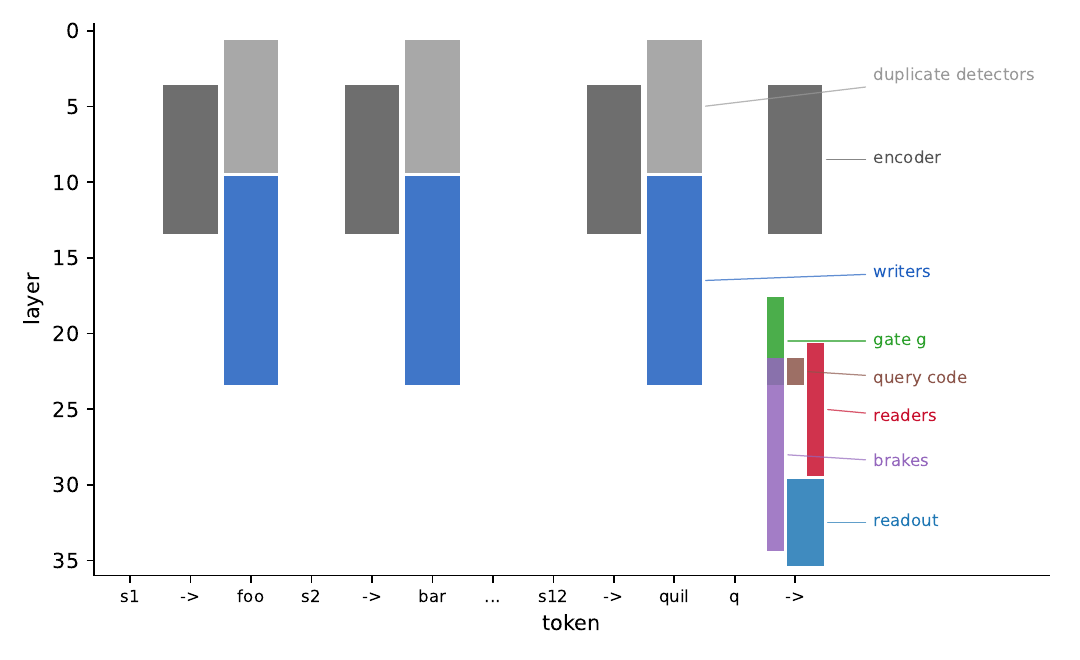}
\caption{One twelve-demonstration prompt, stage by stage. Columns are tokens (sentences $s_i$, arrows, labels and the query $q$) and rows are layers of Qwen3-4B. Each coloured block marks the tokens and layers where a stage acts: the duplicate detectors and the encoder early on, the writers into the label tokens in the middle layers, and then, at the query arrow, the query code, the gate, the readers, the brakes and the readout band. Later sections give the intervention behind each block.}
\label{fig:walkthrough}
\end{figure}

\begin{figure}[t]\centering
\includegraphics[width=\linewidth]{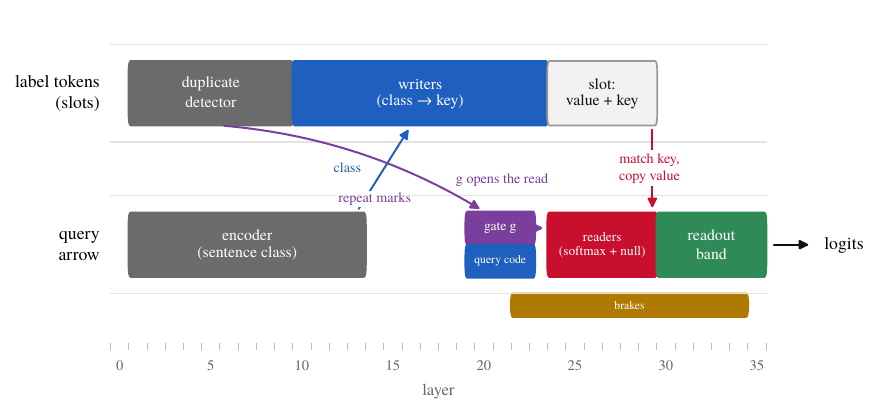}
\caption{The same circuit laid out along the layers of Qwen3-4B-Base. The top lane shows what happens at the label tokens: duplicate-token heads mark a repeated label, and writers deposit the sentence's class into the label's residual stream, which makes it a slot with a value and a key. The bottom lane shows what happens at the query arrow: the encoder computes the query's class, the pooled repeat signal $g$ lowers the readers' sink logit, the readers match the query's class against the slot keys and copy the matching values, and the readout band and late MLPs write the logits while the brakes write against the last label.}
\label{fig:lanes}
\end{figure}

This paper gives such an account for two-class label retrieval and function selection, on invented-word and natural labels, with up to twelve demonstrations, in three base models of three to four billion parameters: Qwen3-4B, SmolLM3-3B and Llama-3.2-3B. In the field's terms we describe a circuit and the algorithm it implements. Figure~\ref{fig:algorithm} draws the circuit, Figure~\ref{fig:lanes} lays it out along the layers, and Figure~\ref{fig:walkthrough} places it on a single prompt.

In brief, the model turns its demonstrations into a key--value memory inside the prompt. Each label token becomes a slot whose value is the identity of the label and whose key is the class of the sentence it followed. The key is deposited by about twenty attention heads in the middle layers, which read the sentence and write its class into the label token's residual stream (K-composition in the sense of \citealt{elhage2021framework}). When the model reaches the query it computes the class of the new sentence, and about ninety heads in the upper layers compare that class with the slot keys through an ordinary softmax over positions and copy out the labels whose keys match. These readers are, in effect, induction heads keyed on a class.

The read is gated. Early in the network, duplicate-token heads register whether any label has appeared twice, and that information is pooled into a single direction in the residual stream at the query. The direction lowers the logit that the readers assign to the attention sink at position zero. When no label repeats, the sink wins the softmax and the readers return nothing in particular. We call this outcome the null action. It follows from the softmax itself, since attention has to go somewhere, and the sink is where these heads put the attention they do not spend on the slots. Finally, a population of about two dozen heads writes against whichever label the model saw most recently. These are negative heads in the sense of \citet{wang2022ioi} and \citet{mcdougall2023copysuppression}. We call them brakes, because removing them nearly triples the margin and leaves the model biased toward the most recent label.

Every step of this account rests on an intervention with a matched control. We build pairs of prompts that are identical token for token except in one span, where we change one sentence, one label or the query, and we move activations from one member of the pair into the other, head by head, group by group or position by position. We knock out heads and compare against random heads from the same layers. We steer the residual stream along directions at the strength the model itself uses and compare against random directions of the same norm. Every head set chosen on one draw of prompts is confirmed on a fresh draw. Where the model concentrates a function in a few heads we name them. Where it spreads a function over many, we describe the population by the direction it reads or writes, and show that this direction matters more than a random subspace of the same dimension. We also account for the answer exactly. On each prompt, the contributions of all heads and MLPs to the answer logit add up to the model's own logit to within 0.05 nats.

Two results surprised us. The first is that the read follows a simple law. The attention each reader pays to the matching slots is described by a softmax expression in the number of matching slots, the total number of slots and the null logit. It has three constants per head, of which two are identifiable, because two of them enter only through their sum. Constants fitted on twelve-demonstration prompts predict a regime with one slot per class that was never used in fitting, in both Qwen and Llama. The same constants fail on longer prompts. When we looked at why, we found that each head's attention to the sink, and its total attention to the matching slots, are roughly fixed properties of the head, so adding matching slots divides a fixed budget among them.

The second surprise concerns why so many of the components are populations. A single reader already separates the two classes in its attention pattern, so the model is not spreading the work because individual heads lack the signal. What individual heads lack is output gain. No head contributes more than 0.2 nats of logit directly, so a 1.3-nat answer is a sum over at least seven heads. (A 1.3-nat change between two prompts needs at least four, since a head's contribution can change sign.)

Sections~\ref{sec:slot} to~\ref{sec:accounting} follow the computation from the slot, through the read (Section~\ref{sec:read}), the gate (Section~\ref{sec:gate}) and the query code (Section~\ref{sec:query}), to the readers and brakes (Section~\ref{sec:populations}) and the final logits. Section~\ref{sec:replication} asks which parts of the account any transformer would need and which are left to training, and tests the account in two further models and across training checkpoints. Sections~\ref{sec:transfer} and~\ref{sec:scale} carry the circuit to natural concepts and word labels, to four model sizes and to an instruction-tuned model, and Section~\ref{sec:build} builds it twice, once by hand and once by training a small transformer from scratch. After related work (Section~\ref{sec:related}), Section~\ref{sec:discussion} sets out what the account does not cover, including the predictions that did not hold, and what we think it implies more broadly.

\section{Setup}\label{sec:setup}
\paragraph{Prompts.} Every prompt in the paper is a list of demonstrations, one per line, written \texttt{sentence -> label}, followed by a query sentence and an arrow. The model's next token is its answer. We use four prompt sets, which we call batteries. The \emph{concept} battery has six demonstrations, three per class, on sentence axes such as topic (animals against food), sentiment, language (English against Spanish) and question against statement, with invented-word labels (\texttt{foo}/\texttt{bar}, \texttt{quil}/\texttt{zorp}) and natural ones. The \emph{concept12} battery is the same with twelve demonstrations, six per class. The \emph{slotgate} battery has four demonstrations and comes in token-identical pairs. In one member of each pair every label appears twice, and in the other every label appears once, so the two differ only in whether labels repeat. The \emph{manyclass} battery has two to six classes with one or more demonstrations each and arbitrary single-token labels. Appendix~\ref{app:prompts} shows one verbatim prompt from each battery, and Appendix~\ref{app:setup} gives further details. Every label except \texttt{zorp} is a single token in all three tokenizers; for a two-token label, the first token carries the key.

\paragraph{The margin.} We score a prompt by its margin, the log-probability of the correct label minus that of the other label, each scored over its full token sequence. A positive margin means the model prefers the right answer, and a margin of 2 nats means it prefers it by a factor of about seven. The effect of an intervention is the change in margin it causes, paired prompt by prompt with the intact run, and every effect is reported with a 95\% interval from a paired bootstrap over prompts (4000 resamples). When we say that knocking out a head set ``costs $-0.74$ nats'', we mean that it lowers the margin by that much on average.

\paragraph{Twin prompts.} Most of our interventions use twins: two prompts that are identical token for token except inside one span. We use five kinds. The \emph{class-flipped} twin replaces one demonstration's sentence with a sentence of the other class. The \emph{same-class} twin replaces it with a different sentence of the same class and the same length; its answer should not change, and it serves as a control. The \emph{label exchange} swaps \texttt{foo} and \texttt{bar} everywhere. The \emph{other-class query} replaces the query sentence with one of the other class. The \emph{unique-label rewrite} replaces each repeated label with a fresh one. Because twins line up token for token, we can take any activation from one and place it into the other at the same position, and ask how much of the difference between the twins that one activation carries. We call this a handoff.

\paragraph{Interventions.} Alongside handoffs we use three other operations. To \emph{knock out} a head we mean-ablate it, replacing its output by its mean over the battery; this removes what the head writes without setting it to zero. To \emph{steer}, we add a direction to the residual stream at the query arrow at the dose the model itself uses, taken to be the difference in projection between twins, and compare the result against a random direction of the same norm. To \emph{hold} a component, we fix it at its intact value while a handoff runs upstream of it, which tells us whether the upstream effect passes through that component. All of these are forward hooks on the HuggingFace implementations of the models, run in bfloat16 on one H100. We also record per-head attention patterns at chosen layers, the direct contribution of every head and MLP to the answer logit, and projections of the residual stream onto fitted directions. Table~\ref{tab:terms} collects the terms we use for the parts of the circuit and for these operations.

\paragraph{Controls.} Every named head set is compared with random sets of the same size drawn from the same layers, and every steering direction with a random direction of the same norm. Every set selected on one draw of prompts is confirmed on a fresh draw from the same pools with a different seed, and the numbers we lead with come from the fresh draw.

\paragraph{Models.} We study Qwen3-4B-Base (36 layers, 32 query heads, 8 key--value groups, QK-norm), SmolLM3-3B-Base (36 layers, 16 heads, 4 groups) and Llama-3.2-3B (28 layers, 24 heads, 8 groups), and we use intermediate checkpoints of OLMo-2 1B and 7B to study how the circuit emerges during training. We write L$\ell$h$h$ for head $h$ of layer $\ell$, both counted from zero, and $\ell$:$h$ for the same head inside lists.

\begin{table}[t]\centering\small
\caption{Terms used throughout the paper.}\label{tab:terms}
\begin{tabular}{@{}lp{0.74\linewidth}@{}}
\toprule
Term & Meaning \\
\midrule
slot & the residual stream at a label token, which holds a key and a value \\
key & the class of the sentence the label followed; written into the slot by writer heads, read through the readers' $W_K$ \\
value & the identity of the label; read through the readers' $W_V$ \\
writer heads & about twenty middle-layer heads that copy a sentence's class into its label token \\
reader heads & about ninety upper-layer heads that attend from the query arrow to slots whose key matches the query's class, and copy their values \\
query code & the representation of the query's class that the readers match against the keys \\
gate $g$ & a direction at the query arrow that signals that some label has repeated, and opens the read \\
null action & the readers putting their attention on the sink at position zero and copying nothing \\
brake heads & about two dozen heads that write against the most recently seen label \\
copier heads & heads in the last six layers that turn what the readers collect into logits \\
\midrule
margin & log-probability of the correct label minus that of the other label, in nats \\
twin & a prompt identical to another except inside one span \\
knockout & mean-ablation of a head's output \\
handoff & patching an activation from a twin into the intact run \\
steering & adding a direction to the residual stream at the dose the model itself uses \\
hold & fixing a downstream component at its intact value while a handoff runs \\
\bottomrule
\end{tabular}
\end{table}

\section{The memory: a key and a value at each label token}\label{sec:slot}
\begin{figure}[t]\centering
\includegraphics[width=\linewidth]{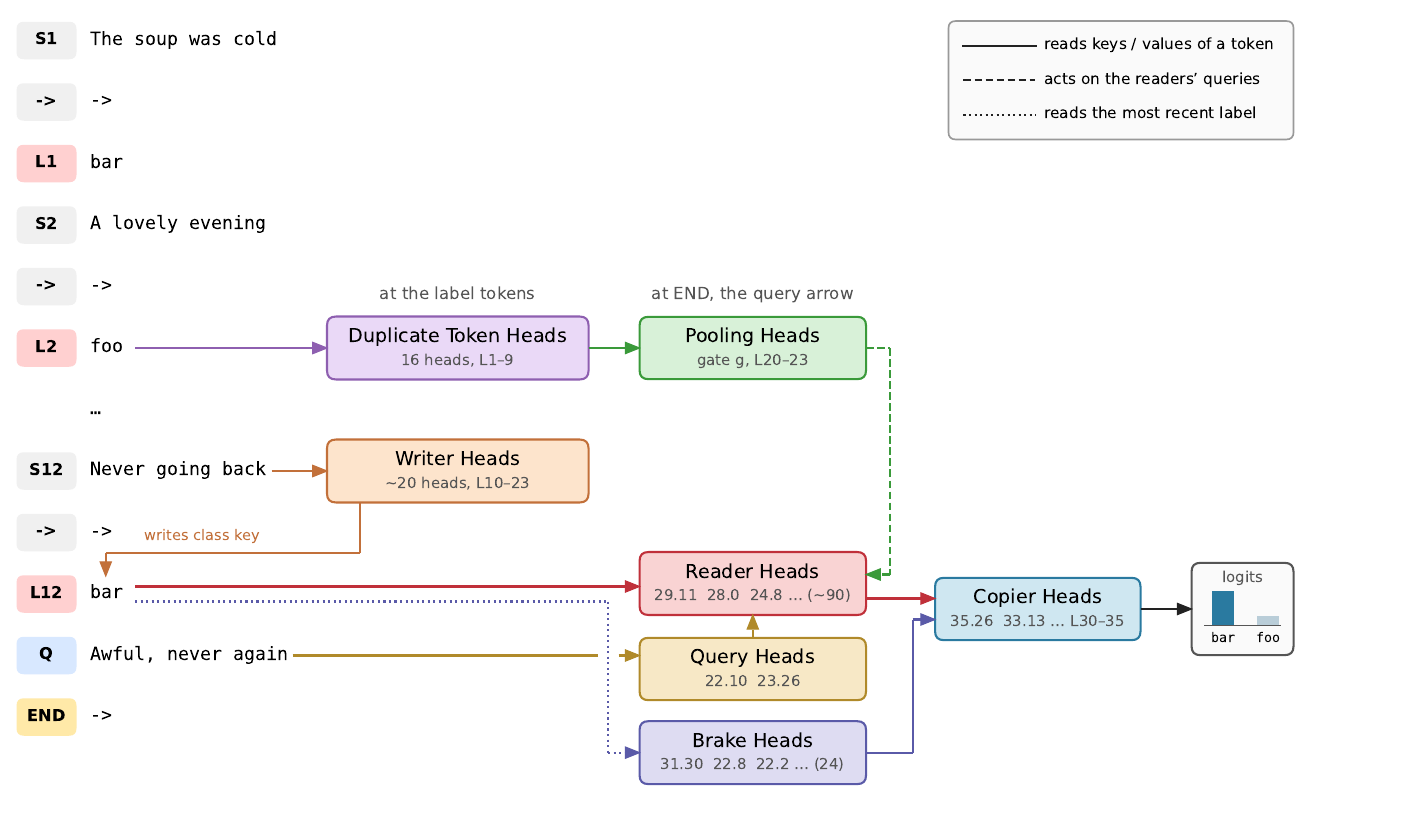}
\caption{The circuit for few-shot in-context learning in Qwen3-4B-Base. Prompt tokens run down the left (sentences $S_i$, labels $L_i$, the query $Q$ and its arrow END). Boxes are classes of attention heads, with layer.head numbers beneath. Writer heads give each label token a key (the class of its sentence) alongside its value (the label). At END, reader heads match the query's class against those keys and copy the matching labels, copier heads write the logits, and brake heads write against the most recent label. The gate $g$, assembled by the pooling heads, opens the read by lowering the readers' attention-sink logit (dashed). Sections~\ref{sec:slot}--\ref{sec:accounting} give the intervention behind each arrow.}
\label{fig:algorithm}
\end{figure}

\begin{figure}[t]\centering
\includegraphics[width=\linewidth]{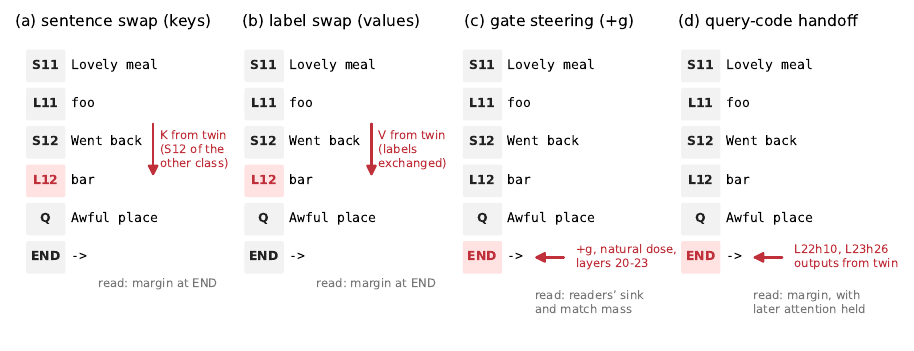}
\caption{The four main interventions, each drawn on the end of a 12-shot prompt. The red component is replaced or steered, and the output is read at END. (a)~Sentence swap: the keys at a label token come from the twin in which that demonstration's sentence belongs to the other class, so the slot's key no longer matches its value. (b)~Label swap: the values at the label token come from the twin with the labels exchanged. (c)~Gate steering: the model's own gate direction $g$ is added at END at its natural magnitude, before the reader layers, and we measure the readers' attention to the sink and to the matching slots. (d)~Query-code handoff: the outputs of the two query heads at END come from the twin whose query is of the other class; later attention outputs can be held fixed to test whether the effect passes through them.}
\label{fig:interventions}
\end{figure}

\begin{figure}[t]\centering
\fig{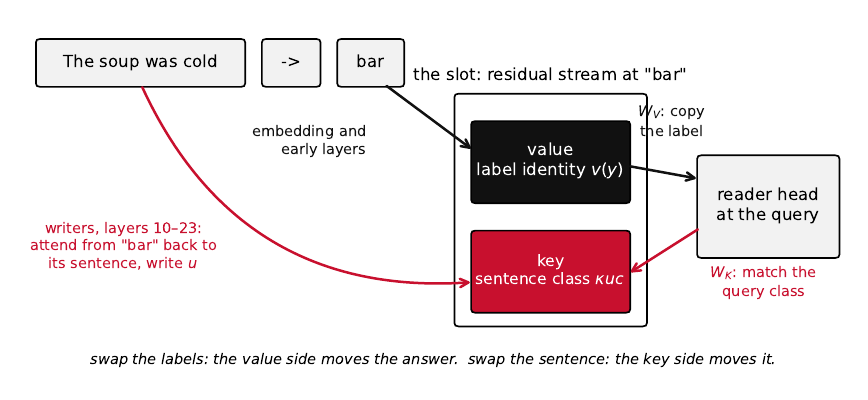}
\caption{Anatomy of a slot. The residual stream at a label token holds two things, which are read through two different matrices. The label's identity comes from the token's own embedding and the early layers, and is read by the readers' $W_V$. The class of the preceding sentence is written in by middle-layer heads that attend back to the sentence, and is read by the readers' $W_K$ as the key that the query's class is matched against.}
\label{fig:slotanatomy}
\end{figure}

By the time the model reaches the query, it needs two facts about each demonstration: which class its sentence belonged to, and which label followed it. Both end up in the same place, the residual stream of the label token. We call that residual a slot, because it behaves like one cell of a key--value memory (Figure~\ref{fig:slotanatomy}). Its value is the identity of the label, and its key is the class of the sentence that came before it. The value and the key are read on different sides of attention, and two experiments, sketched in Figure~\ref{fig:interventions}a--b, pull them apart.

\paragraph{The value: exchanging the labels.} Take a prompt and swap \texttt{foo} and \texttt{bar} throughout, leaving every sentence in place. The right answer to the query is now the other label, and the model's margin changes sign. To find which activations carry the flip, we patch the keys and values at the label tokens from the swapped prompt into the intact run. Patching them at every layer reproduces the flip exactly ($-7.18$ nats in Qwen, $-5.55$ in SmolLM3, $-4.92$ in Llama), but that only checks the harness and says nothing about the model. Two prompts that differ only at the label tokens have identical inputs everywhere else, so replacing the label tokens' keys and values at every layer makes every other position compute exactly what it computes in the swapped prompt. This is true of any transformer, by induction over layers (Appendix~\ref{app:induction}).

\begin{figure}[t]\centering
\fig{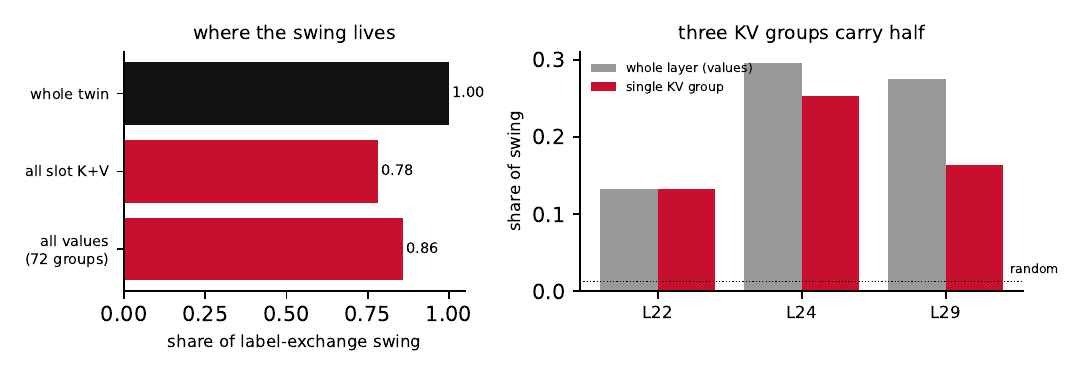}
\caption{Three key--value groups carry half of the label's value. Patching the values of all 72 KV groups of layers 21--29 at the label tokens, from a prompt with the labels exchanged, carries 86\% of the exchange's effect on the margin (left). By layer, the effect sits in layers 22, 24 and 29 (right, grey), and within each of these layers one KV group carries most of it (red). Group 24.5 alone carries a quarter, and groups 24.5, 29.2 and 22.5 together carry about half. Random head sets of the same size move the margin by 0.03 or less (dotted). Qwen3-4B, 12-shot invented-label prompts, 72 prompts.}
\label{fig:transport}
\end{figure}

The informative patches are the restricted ones. In SmolLM3, the keys and values at the label tokens in the reader layers 21--29 alone carry $-2.94$ of the $-5.55$. In Qwen, the values of the 72 key--value groups of layers 21--29, at the label tokens, carry 86\% of the exchange. The values of the eight sharpest readers carry $-2.16$, while their keys carry $+0.56$, which goes the wrong way. The label's identity therefore sits on the value side of the slot, and it is read in the upper layers. Within the value side it is concentrated. Patching the values of single key--value groups shows that three of the 72 groups in layers 21--29 carry about half of the exchange's effect, and that one group in layer 24 carries a quarter (Figure~\ref{fig:transport}).

\begin{figure}[t]\centering
\fig{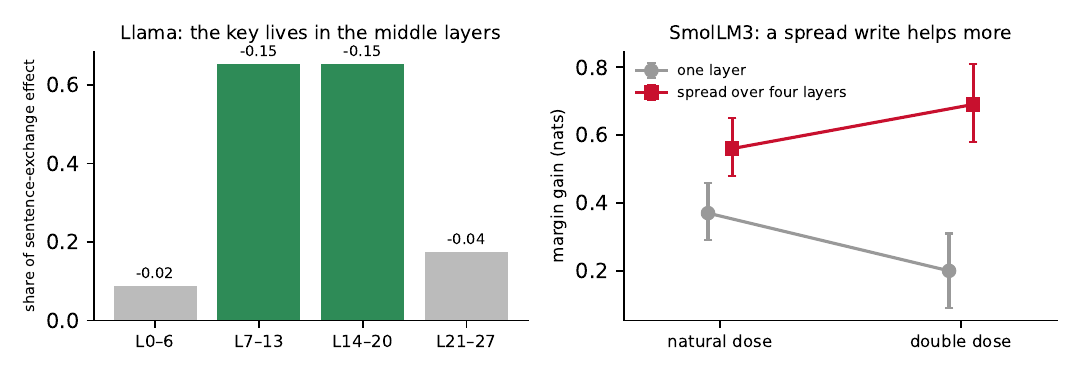}
\caption{The key is written in the middle layers, and the write has to be spread out. (a)~In Llama-3.2-3B, patching the label-token keys and values from the sentence-exchanged prompt one layer band at a time: the two middle bands carry the effect, and the first and last carry almost none. (b)~In SmolLM3, adding a fixed total dose of the class direction $u$ to the label tokens at one layer helps less than spreading it over four layers, and falls back at double dose while the spread write keeps climbing (95\% intervals; 144 prompts).}
\label{fig:writers}
\end{figure}

\paragraph{The key: changing a sentence.} The second experiment leaves the labels alone and changes a sentence instead. We replace one demonstration's sentence with a sentence of the other class, so that \texttt{foo} now follows a complaint where it used to follow a compliment. This also moves the answer, by $-2.87$ nats on the six-demonstration battery, but now the effect runs through the keys. Patching the changed prompt's keys at the label tokens in layers 21--29 carries $-2.35$ of it, and the eight sharpest readers' keys carry $-0.55$ while their values carry $+0.10$. In Llama the same experiment places the key in the label tokens' keys and values in the middle layers only. Layers 7--13 and 14--20 each carry $-0.15$ of a $-0.23$ twin effect, while layers 0--6 and 21--27 carry $-0.02$ and $-0.04$ (Figure~\ref{fig:writers}a). The early layers have not yet written the key, and by the late layers it has already been read.

\begin{figure}[t]\centering
\fig{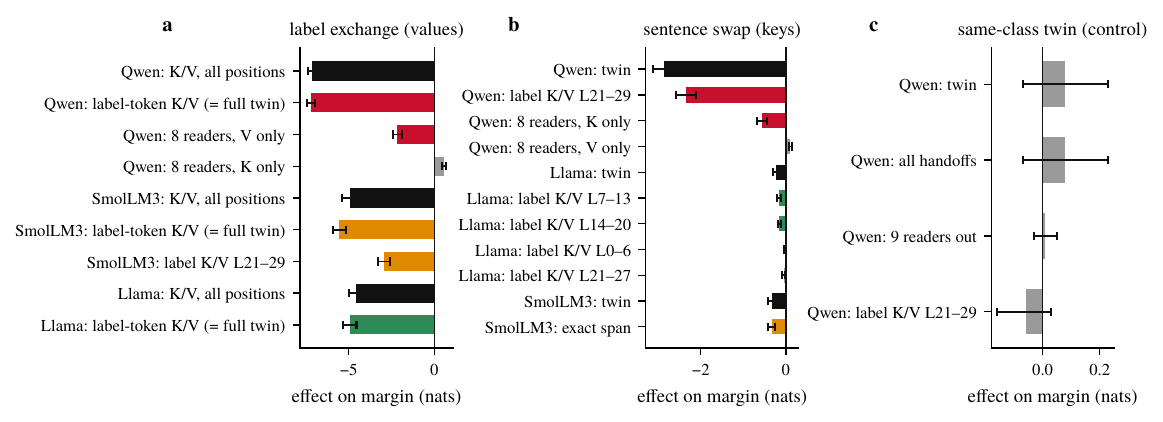}
\caption{The slot holds a value and a key on different sides of attention. (a)~Exchanging the labels: in all three models the effect runs through the values at the label tokens. (b)~Changing a sentence while keeping its label: the effect runs through the keys at the label tokens, and only in the middle layers. (c)~Handoffs from a same-class sentence of the same length move the answer by at most 0.06 nats. Bars are paired effects on the margin with 95\% bootstrap intervals; grey bars are size-matched random controls.}
\label{fig:slot}
\end{figure}

A control confirms that this second experiment measures the class and not the particular sentence. Replacing a demonstration's sentence with a different sentence of the same class and the same token length moves the answer by $+0.08$ nats, and every handoff from that twin stays within $\pm0.06$. Whatever else the slot stores about its sentence, only the class reaches the answer. Figure~\ref{fig:slot} summarizes the three experiments.

\paragraph{Writing the key.} The label token cannot see its sentence directly, so attention has to carry the class to it. About twenty heads in layers 10--23 do this. Each attends from the label token back to the words of its own sentence and writes the class direction into the label's residual stream. No one of them carries more than 5\% of the effect, the top twenty together carry 64--92\%, and knocking out six named writers does no more than knocking out six random heads from the same layers. What the writers write is a single direction $u$ per task, the same for both labels (held-out sign agreement 0.95--1.00), so a two-class key needs only one dimension.

The write also has to be spread across depth: adding a given total dose of $u$ to the label tokens at one layer helps less than spreading the same dose over four layers. In SmolLM3, on the 144 prompts from the axes the model does not bind on its own (with the query's class steered as well), a write at layer 16 alone gains $+0.37$ \ci{+0.29}{+0.46}, and the same dose split over layers 12, 14, 16 and 18 gains $+0.56$ \ci{+0.48}{+0.65}. At double dose the single-layer write falls back to $+0.20$ \ci{+0.09}{+0.31}, while the spread write keeps climbing, to $+0.69$ \ci{+0.58}{+0.81} (Figure~\ref{fig:writers}b). We take this as a sign that the writers are contextualisation heads. Each layer's write is read and reshaped by the next, and a dose dropped into a single layer does not get the same treatment.

\paragraph{The slot as a bottleneck.} Some features never bind, however many demonstrations the model sees. Word parity, the third letter of a word and rules about digits are examples, and none of them is represented linearly at the arrow. If we write $u$ into the slots of such a prompt by hand, accuracy rises from 50 to 71\%, while key-only, query-only and random directions stay within $\pm0.1$ (Appendix~\ref{app:encoder}). So when the key is present the rest of the circuit can use it, and when it is missing, nothing downstream makes up for it. We turn next to how the slots are read.

\section{Reading the memory: reader heads and a softmax law}\label{sec:read}
The lookup is done by attention heads in the upper layers, which we call readers. Each reader performs an ordinary softmax over the positions of the prompt, and the attention sink at position zero takes part in that softmax like any other position. Because the read is a softmax, the attention a reader pays to the matching slots should follow a simple expression with three constants per head, two of them separately identifiable. We can fit that expression in one regime and test it in another.

\paragraph{Notation.} Write $h_q$ for the residual stream at the query arrow at the reader's layer, and $r_j$ for the residual stream at the $j$-th label token. The head forms $q = \mathrm{RoPE}(\gamma_q \odot \mathrm{norm}(W_Q h_q))$ and $k_j = \mathrm{RoPE}(\gamma_k \odot \mathrm{norm}(W_K r_j))$, where norm is the per-head RMS normalization that Qwen applies and $\gamma$ is its gain; Llama and SmolLM3 omit the normalization. The logit for position $j$ is $\ell_j = q^\top k_j / \sqrt{d}$, and the attention weight is $\mathrm{softmax}_j(\ell)$ over all positions up to the query. We write $N$ for the number of slots, $n_m$ for the number whose class matches the query, $L$ for the logit of a matching slot and $L_{\mathrm{sink}}$ for the logit of position zero, and we choose the origin so that the non-matching slots have logit zero.

\begin{proposition}[Matching mass]\label{prop:law}
Suppose the matching slots share the logit $L$, the non-matching slots the logit $0$, the sink the logit $L_{\mathrm{sink}}$, and the remaining positions together contribute $n_{\mathrm{rest}}$ units of unnormalized mass. Then the attention mass on the matching slots is
\begin{equation}
m \;=\; \frac{n_m e^{L}}{\,n_m e^{L} + (N-n_m) + n_{\mathrm{rest}} + e^{L_{\mathrm{sink}}}\,}.
\label{eq:law}
\end{equation}
\end{proposition}
\begin{proof}
The softmax denominator is the sum of $e^{\ell_j}$ over positions; group the terms by type and divide the matching slots' share by the total.
\end{proof}

As algebra the proposition is trivial. Its content is the empirical claim that three numbers per head (two of them identifiable, since $n_{\mathrm{rest}}$ and $e^{L_{\mathrm{sink}}}$ enter only through their sum), fitted once, describe how the head behaves as the prompt changes. We tested this by varying $n_m$ and $N$. On twelve-demonstration prompts with two, three, four or six classes (so that $n_m$ is six, four, three or two), Equation~\ref{eq:law} with three fitted constants reproduces the measured matching mass to within 0.1 for 127 of the 140 Qwen heads that put at least 0.03 of their attention on slots. In Llama it does so for 124 of 125 such heads, with a median maximum error of 0.02. The sharpest readers put 0.6--0.8 of their attention on the matching slots when there are two classes and about 0.45 when there are six, and the diffuse ones put a few hundredths.

\begin{figure}[t]\centering
\fig{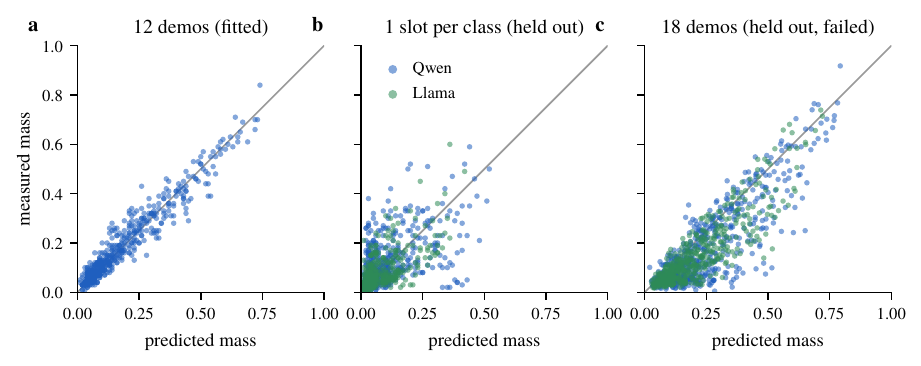}
\caption{Within a regime the read follows a softmax law, and the law transfers to fewer slots per class but not to longer prompts. Each point is one head at one slot count, showing the measured matching-slot mass against the prediction of Equation~\ref{eq:law} from twelve-demonstration constants. (a)~Qwen, in the fitted regime. (b)~Qwen and Llama, one slot per class (held out). (c)~Qwen and Llama, eighteen demonstrations (held out; the prediction fails). Sharp readers sit on the line and diffuse heads fall below it.}
\label{fig:law}
\end{figure}

A good fit within one regime is weak evidence, so we also asked the law to predict a regime it had not seen. Keeping each head's three constants and setting $n_m = 1$, the law predicts the battery with one slot per class with a median error of 0.07 in Qwen (69\% of heads within 0.1) and 0.04 in Llama (83\%). Figure~\ref{fig:law} plots measured against predicted mass for both models.

Predicting in the other direction is where the law breaks down. We computed every head's predicted mass on eighteen-demonstration prompts, which have three to nine slots per class, from the twelve-demonstration constants, and set ourselves the criterion of a median error below 0.07. The observed medians were 0.15 in Qwen and 0.13 in Llama, and 92--94\% of heads were over-predicted. The attention patterns show why. A head's null mass, its attention to the sink, is a constant of the head. The sharpest Qwen readers put 0.34--0.36 of their attention on the sink at every slot count and at both prompt lengths, and Llama's put 0.42--0.45. A head's total attention to the matching slots is nearly constant too. \head{29}{11}, for example, puts 0.18 on the matching slots when there are six of them and 0.19 when there are nine, so the extra matching slots divide a fixed class budget among themselves. Within one prompt length the growth from two to six slots per class is real, which is why the fit worked, but it saturates, and the fit absorbed the saturation into $n_{\mathrm{rest}}$ and $L$.

Equation~\ref{eq:law} is therefore a law for a fixed prompt length, and it transfers to fewer slots per class but not to longer prompts. The failure also established two facts that the rest of the paper relies on: the gate acts on a null mass that is fixed for each head, and because each head's class budget is fixed, the readers have to act together to produce a large answer.

\section{A repeat signal gates the read}\label{sec:gate}
\begin{figure}[t]\centering
\fig{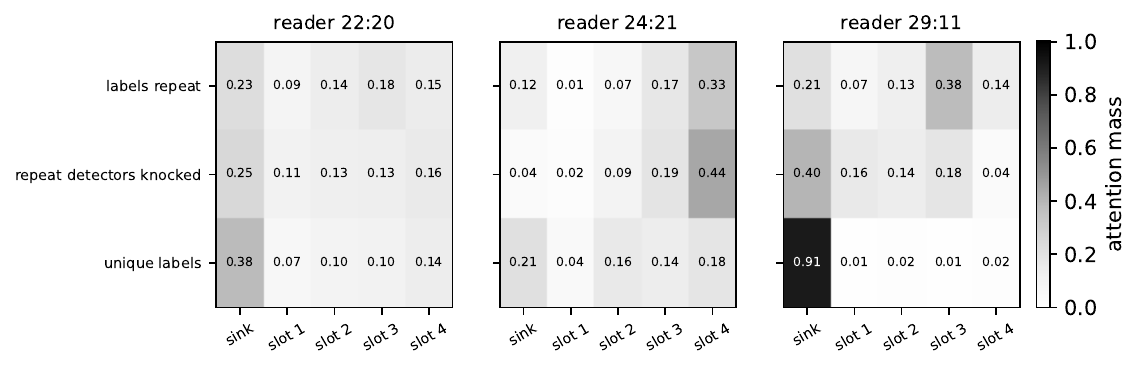}
\caption{What a reader looks at. Attention from the query arrow to the first position (the sink) and to the four label slots, for three Qwen3-4B readers, averaged over 30 four-demonstration prompts. Top row: two labels, each used twice; the matching slot (slot 3 or 4, depending on the prompt) draws the attention. Middle row: the same prompts with the sixteen repeat-detector heads mean-ablated. \head{29}{11} moves back toward the sink (0.21 to 0.40), while \head{24}{21} and \head{22}{20} do not, so the detectors' signal reaches some readers and not others. Bottom row: four distinct labels, none repeated; \head{29}{11} puts 0.91 of its attention on the sink and reads nothing.}
\label{fig:readerrows}
\end{figure}

A memory that is always read would be a liability. Most text contains no demonstrations, and a model that copied a label after every arrow would usually be wrong. The read should switch on only when there is something to read, and in these models the switch is a single direction in the residual stream at the query. It is computed from whether any label has appeared twice, and it works by lowering the logit that the readers assign to the attention sink. Figure~\ref{fig:readerrows} shows three readers in three states: reading when a label repeats, partly back on the sink when the repeat detectors are removed, and on the sink throughout when no label repeats.

What first pointed us to a gate was the way accuracy depends on repetition. With one demonstration per class the model retrieves the right label about 20\% of the time, with two per class 60--80\% of the time, and with six, 96\%. The requirement applies to the prompt as a whole. A label that appears only once is retrieved correctly as long as the other labels in the prompt repeat, so something in the statistics of the whole prompt opens the read.

\paragraph{The gate is one direction.} We call this direction $g$ and fit it on the residual stream at the query arrow in layer 23. There it separates prompts whose labels repeat from their token-identical twins whose labels are all distinct, with held-out sign agreement of 1.00, and it is orthogonal to the key direction $u$. Prompts with repeated labels project onto it at about $+18$, and their unique-label twins at about $-21$.

\begin{figure}[t]\centering
\fig{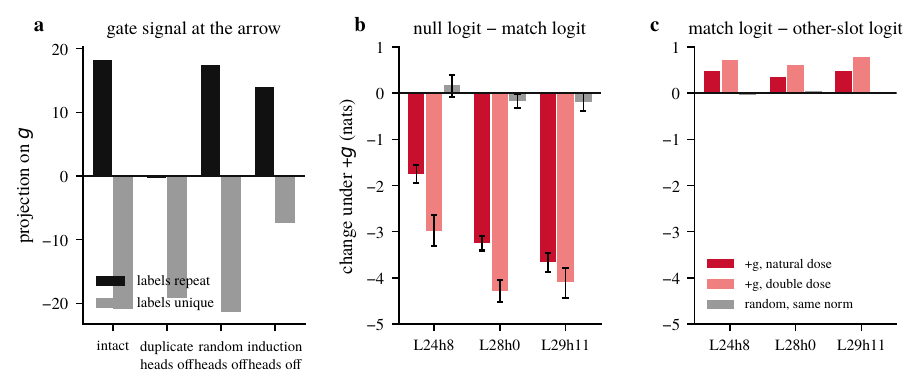}
\caption{The gate lowers the null logit and moves matching far less. (a)~Projection of the arrow residual onto $g$ for prompts with repeated and with unique labels, intact and with the sixteen duplicate-token heads ablated, with random and induction-scored sets as controls. (b)~Change in each reader's sink-minus-match logit when $g$ is added at natural and double dose; same-norm random directions in grey. (c)~Change in the same readers' match-minus-other logit.}
\label{fig:gate}
\end{figure}

The signal reaches the arrow through flat pooling heads in layers 20--23, which attend evenly over the slots. Patching their values from the twin carries 93--100\% of the projection, and patching their keys carries 3--10\%. The slots carry the signal in the first place because of sixteen duplicate-token heads in layers 1--9, which attend from each label token to earlier copies of the same token. Mean-ablating these sixteen heads drives the projection on recurring-label prompts from $+18.3$ to $-0.3$, about half of the way to the unique-label value of $-21$. Sixteen random heads move it by at most 0.7, and the sixteen heads with the highest induction scores move it by 4.2 (Figure~\ref{fig:gate}a). The other half of the separation comes from elsewhere, and we do not know where. The duplicate-token detector is the first half of the classic induction circuit, and here it has a second job, which is to report that there is something to copy.

\paragraph{The gate lowers the null logit.} The readers' softmax includes the sink at position zero, and a reader that puts most of its attention there copies nothing. Adding $g$ to the residual stream at the arrow of a unique-label prompt, at the dose the model itself uses, lowers each reader's sink logit relative to its matching-slot logit, by 1.75 nats at \head{24}{8}, 3.25 at \head{28}{0} and 3.65 at \head{29}{11}. Across the distributed readers the drop is 0.74 at the natural dose and 1.41 at double dose, and a random direction of the same norm moves it by at most 0.2 (Figure~\ref{fig:gate}b). The gap between the matching and non-matching slot logits also widens, by $+0.34$ to $+0.48$ nats (a factor of about 1.5 in the attention ratio), roughly a seventh of the effect on the sink (Figure~\ref{fig:gate}c). Patching in the whole recurring-label twin moves the sink logit by the same 1.5--3.9 nats and leaves that gap unchanged or smaller. The gate acts mainly on the null logit, and its smaller effect on matching is not zero.

The weights tell the same story. $g$ projects onto the sink-versus-slots query functionals of the 89 upper-layer slot-reading heads with the sign that releases the read. The population cosine is $-0.03$, against a random $|\cos|$ of 0.015, and the clearest individual heads reach $-0.09$ to $-0.11$. It does not project onto the same heads' match functionals ($|\cos| \le 0.02$). It carries 10--11\% of its energy in the span of these heads' functionals, against 3.3\% for a random subspace of the same dimension. The sign and the selectivity are therefore properties of the weights, while the magnitude is spread over more heads than our census of slot-reading heads contains.

\begin{corollary}[Null action]\label{cor:null}
Under the conditions of Proposition~\ref{prop:law}, if $e^{L_{\mathrm{sink}}} > n_m e^{L}$ the reader returns more of the sink's value than of the matching slots', and a signal present only at the query position can change this only through a term that enters $L_{\mathrm{sink}} - L$, that is, through the reader's query. (A signal written at the slot positions would enter through the keys instead, which is not how the gate acts.)
\end{corollary}

Corollary~\ref{cor:null} explains why the gate acts where it does. A softmax must put its mass somewhere, and for these heads position zero is the sink, the place where they put the attention they do not spend on the slots. (Its value is not inert, as Section~\ref{sec:arrows} shows, but the heads use it as a place to put unspent attention.) A reader whose null logit exceeds its match logits returns the sink's value, and the model emits nothing in particular. This is the null action. It is a consequence of the architecture, and the only lever a prompt-level signal has on it is the reader's query, which is where $g$ acts, with the sign the corollary requires.

Llama-3.2-3B's gate differs in strength and in how it is controlled. At four demonstrations 67\% of its unique-label prompts still receive a label, against 10\% in Qwen, and the heads that a generic induction score picks out hold Llama's middle-layer readers on the sink instead of releasing them (Section~\ref{sec:replication}).

\section{Two heads write the query code}\label{sec:query}
The slot keys are one side of the match, and the other side is whatever the readers bring from the query. One might expect the query to carry the same class direction $u$ that the slots hold, so that the match is a dot product of like with like. It turns out that the query carries a code of its own, written by two heads and shaped for the readers that consume it.

The experiment is the query flip. We replace the query sentence with one of the other class and patch single activations from the flipped prompt into the intact run at the arrow. Patching the attention outputs of layers 21--29 carries $-1.10$ \ci{-1.39}{-0.82} of the $-1.33$-nat flip in Qwen, so the readers' layers are where the query's class acts. Patching the outputs of just two heads one layer earlier, \head{22}{10} and \head{23}{26}, carries $-0.53$ \ci{-0.68}{-0.39}, or 40\% of the flip, while any pair of random heads from the same layers carries less than 0.05. These two heads compute the query's class at the arrow.

What they write is not $u$. It has cosine 0.07 with the key direction, and it does not line up with the weight-probe direction of any single reader. It is, however, the direction that moves the reader population's match logits. For each reader, the query-side functional for ``match against the class-$c$ key'' is a vector in residual space, $W_Q^\top$ applied to the mean key difference. The span of these functionals over the readers of layers 24--29 contains 64\% of the pair's write, against 4\% for a random subspace of the same dimension. In other words, the code is written in the readers' own coordinates.

To find which components consume the code, we held downstream components at their intact values while the pair's handoff ran. Holding the attention queries of layers 24--29 removes most of the effect ($-0.14$ remains), holding layers 24--35 removes 93\% of it ($-0.04$ remains), and holding the late MLPs as well leaves $-0.01$. Holding the late MLPs alone makes the effect larger ($-0.66$), which means that in the intact run they push against the query code. Putting this together, two heads write the code at layers 22--23, the attention queries of layers 24--35 read it (the copiers of layers 30--35 take the last quarter), and the final MLPs oppose it.

\begin{figure}[t]\centering
\fig{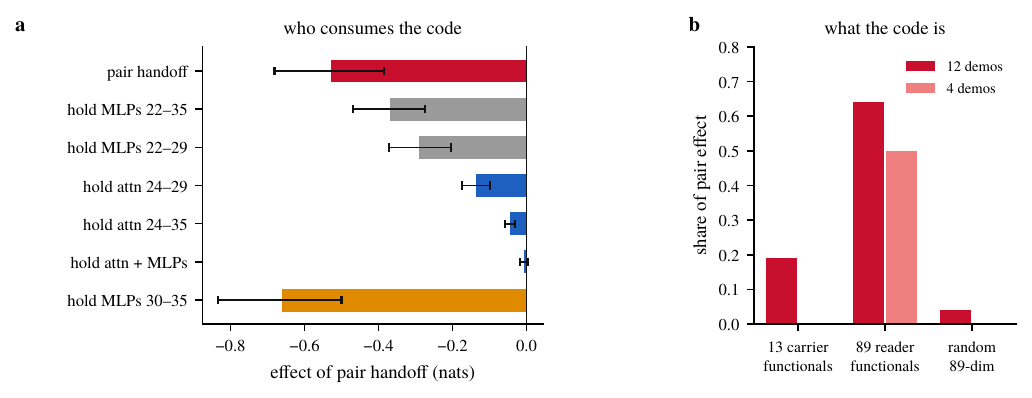}
\caption{Two heads write a query code that the reader population consumes. (a)~Share of the query-flip effect carried by the pair \head{22}{10}+\head{23}{26} and by all attention in layers 21--29, and what remains when downstream attention queries are held at their intact values. (b)~Energy of the pair's write in the span of the readers' match functionals, against random subspaces of the same dimension.}
\label{fig:query}
\end{figure}

How concentrated the query computation is varies by model. SmolLM3 spreads it over more heads, with no pair carrying 40\%, and Llama's query-side effect spans layers 10--27 (band handoffs of $-0.03$, $-0.18$ and $-0.13$ for layers 0--9, 10--17 and 18--27). In all three models, though, the query carries a code that is separate from the key and is consumed by the readers' queries. Figure~\ref{fig:query} shows the consumer map and the functional-span test.

\section{Readers and brakes act as populations}\label{sec:populations}
\begin{figure}[t]\centering
\includegraphics[width=\linewidth]{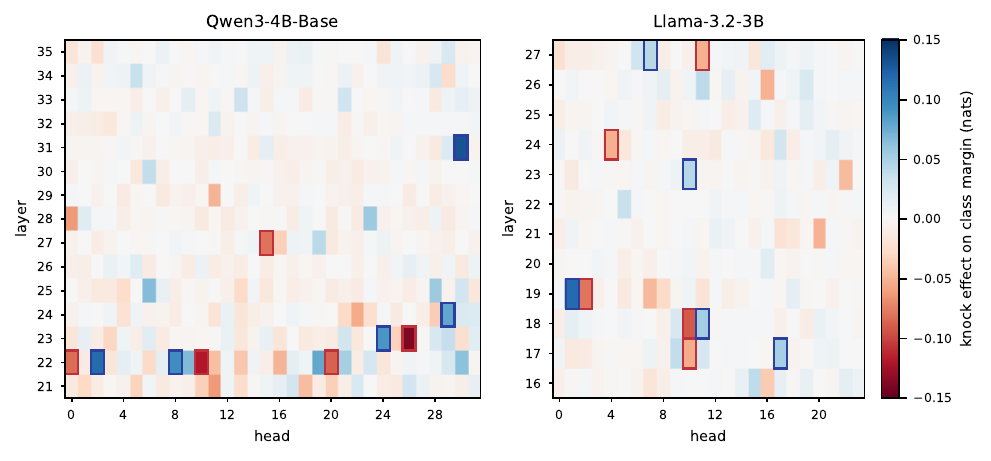}
\caption{No single head carries the answer. Each cell is the effect on the 12-shot class margin of knocking out one head at the query position (mean over 144 prompts from six label families; blue lowers the margin and red raises it; colour clipped at $\pm0.15$ nats), for every head in the reader layers of Qwen3-4B-Base (layers 21--35) and Llama-3.2-3B (layers 16--27). Outlined cells are the five most negative and five most positive heads. The largest single-head effects are $-0.14$ nats (Qwen \head{23}{26}, a query head) and $+0.13$ (Qwen \head{31}{30}, a brake), and $-0.09$ and $+0.12$ in Llama (\head{18}{10}, \head{19}{1}), against a 1.3-nat margin. The sets built from these rankings move the margin by $-0.74$ and $+1.39$ in Qwen and by $-0.56$ and $+1.90$ in Llama when knocked out together (Table~\ref{tab:claims}).}
\label{fig:headgrid}
\end{figure}

There are about ninety readers, and about two dozen heads work against them. Both numbers are large next to the handful of heads that circuit analyses usually name, and in this section we explain why the model uses so many.

A reader, concretely, is an upper-layer head whose attention from the arrow prefers the slots whose class matches the query. Twenty readers chosen on one draw of prompts and knocked out together on a fresh draw cost $-0.74$ \ci{-0.95}{-0.55} nats, while twenty random heads from the same layers cost $+0.02$. No single reader is necessary. The largest single-head knockout among the readers in Qwen costs 0.09 nats, and the two query heads, 23:26 and 22:10, cost 0.14 and 0.12 (Figure~\ref{fig:headgrid}). Against the readers stand the brakes, twenty-four heads in layers 22--34 that write against the label the model has just seen. Knocking them out on a fresh draw raises the margin by $+1.39$ \ci{+0.88}{+1.88}, nearly tripling it, while twenty-four random heads give $+0.22$. The brakes are negative heads, like the negative name movers of the indirect-object circuit and the heads that perform copy suppression. They take a label that is being copied and push the logits the other way.

\begin{figure}[t]\centering
\fig{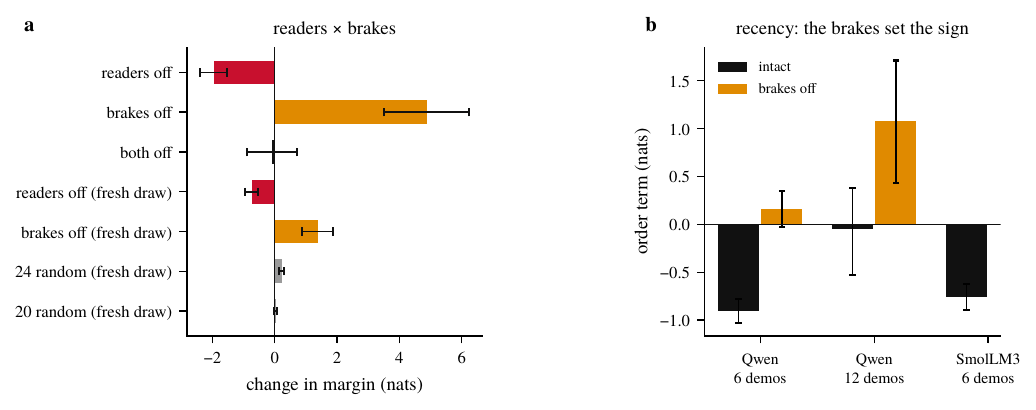}
\caption{Readers and brakes act as populations with opposite signs. (a)~Joint knockout of 20 readers, of 24 brakes, and of both, on a fresh draw, against size-matched random sets (grey). (b)~Order effect: the margin against the position of the last matching demonstration, intact and with the brakes knocked out.}
\label{fig:populations}
\end{figure}

\paragraph{What the brakes do.} The brakes calibrate the model against the order of the demonstrations. Permuting the demonstrations leaves the mean margin unchanged, but on half of the prompts it moves the margin by more than a nat. The readers themselves are biased toward recent slots, with a gradient of 0.5--0.85 nats from the first slot to the last, and with the brakes removed the model follows that gradient and prefers whichever label came last. At six demonstrations the brakes supply the anti-recency term outright, and at twelve they keep the model order-neutral against the readers' pull. Figure~\ref{fig:populations} shows the interaction between readers and brakes and the order effect, and Appendix~\ref{app:order} gives the details.

\paragraph{Why the read needs many heads.} One might think the model spreads the read over ninety heads because no single head can tell the classes apart. The attention patterns rule this out. A single reader's class bias is 0.2--0.4 nats against a per-slot spread of about 0.5, so one head already separates the classes in its own attention pattern, if weakly. What no head has is enough output gain.

\begin{proposition}[Gain bound]\label{prop:gain}
Let head $i$ attend with mass $m_i$ to slots whose values differ by $\Delta v$ between the two labels, with output-value matrix $W_{OV}^{(i)}$, and let $w$ be the final-norm-scaled unembedding direction of the correct minus the alternative label. The head's direct contribution to the margin is at most $m_i\,\|W_{OV}^{(i)}\Delta v\|\,\|w\|/\rho$, where $\rho$ is the RMS scale of the final residual. If the margin $M$ is carried by the heads' direct contributions and every head's bound is at most $\beta$, at least $\lceil M/\beta \rceil$ heads contribute.
\end{proposition}

When we measured it, no head in Qwen wrote more than 0.2 nats of the answer logit directly (Section~\ref{sec:accounting}). A 1.3-nat margin is therefore a sum over at least seven heads, and a 1.3-nat class flip, in which a head's contribution may change sign, a sum over at least four. In practice the model uses 14 heads for half of the flip and 46 for 80\%. It is gain that forces the populations, and a lack of signal in single heads plays no part. We therefore describe the populations by the direction they read and write and test that direction against random subspaces. A search for a single responsible head would be looking for something that the bound says cannot exist.

The same arithmetic gave us a prediction to test. Knocking out the seven largest direct movers in the copier band should remove 0.3--0.5 of the $+0.68$ twelve-demonstration margin. It removed $-0.16$ \ci{-0.56}{+0.24}, which has the predicted sign but is smaller than predicted, with an interval that includes zero. We count this as weak support.

\section{Accounting for the answer}\label{sec:accounting}
So far we have located each step by the change it causes. For a single prompt we can also ask a stricter question: which components wrote the answer's logit, and how much did each one write? The residual stream before the final norm is a sum over components, so the answer logit decomposes exactly.

\subsection{Direct contributions}
\begin{proposition}[Direct decomposition]\label{prop:dla}
Let $x = e + \sum_i a_i + \sum_l m_l$ be the final residual at the arrow, with $e$ the embedding, $a_i$ the head outputs and $m_l$ the MLP outputs, and let $\rho = \mathrm{rms}(x)$. With $w = \gamma \odot W_U^\top(\mathbf 1_{y} - \mathbf 1_{y'})$ for the final-norm gain $\gamma$ and labels $y, y'$, the margin is $\mathrm{logit}(y) - \mathrm{logit}(y') = \big(w^\top e + \sum_i w^\top a_i + \sum_l w^\top m_l\big)/\rho$.
\end{proposition}

Proposition~\ref{prop:dla} is an identity, and we use it as a check. On every prompt, the right-hand side computed from the hooked components agrees with the model's own margin, with a mean error of 0.045 nats in Qwen and 0.048 in SmolLM3. With that check in place, the decomposition tells us who writes the answer. On intact twelve-demonstration prompts the margin is $+0.68$ nats. Attention in layers 30--35 contributes $+0.66$, attention in layers 21--29 contributes $+0.14$, and the late MLPs contribute $-0.11$. Under the exact class flip the answer swings by $-1.33$: the layer 30--35 band moves by $-1.12$, the readers by $-0.36$, and the late MLPs by $+0.25$, in the opposite direction.

\begin{figure}[t]\centering
\fig{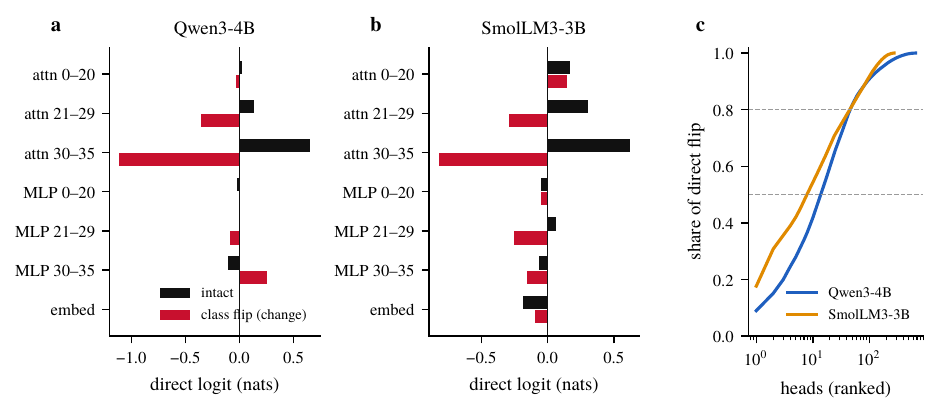}
\caption{Every nat of the answer assigned to a component. (a)~Direct contributions to the margin by layer band, intact and under the class flip, in Qwen and SmolLM3. (b)~Cumulative share of the flip against the number of heads, sorted by contribution.}
\label{fig:accounting}
\end{figure}

\begin{figure}[t]\centering
\fig{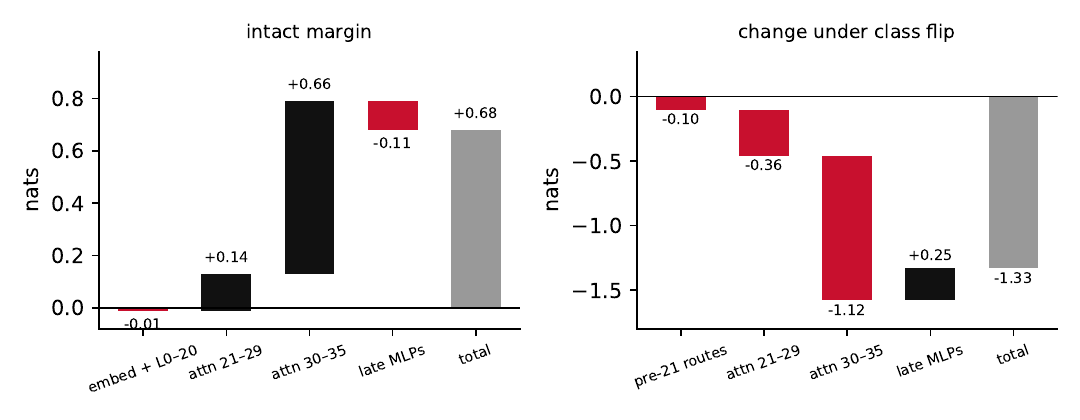}
\caption{The answer as a sum of bands. Left: the intact twelve-demonstration margin in Qwen, built up band by band; the copier band of layers 30--35 writes almost all of it. Right: the change under the exact class flip. The copier band carries most of the swing and the readers a quarter, and the late MLPs push the other way.}
\label{fig:waterfall}
\end{figure}

The readers, then, are the source of the answer, but they write little of it themselves. They collect the label values into the arrow, a band of copier heads in the last six layers reads what they collected and writes the logits, and the final MLPs act as a prior toward the label seen most often and oppose the flip. In Qwen, fourteen heads carry half of the direct swing and 46 carry 80\%; in SmolLM3 the numbers are eight and 45. Figure~\ref{fig:waterfall} shows the decomposition as a waterfall, and Figure~\ref{fig:accounting} shows how the swing is concentrated over heads. The direct accounting credits each nat to the last component that wrote it, while the handoffs of the earlier sections credit it to its source, and between them they close the path from slot to logit. Section~\ref{sec:arrows} shows that the copiers are one of three routes out of the readers, and Appendix~\ref{app:trace} follows a single prompt through every layer and position.

\subsection{Faithfulness, completeness and minimality}\label{sec:faithful}
\citet{wang2022ioi} ask three things of a circuit: that it be faithful, complete and minimal. The exact decomposition above is an accounting identity and does not test faithfulness. The real test is to keep only the named heads and replace the output of every other head with a reference value. We do this at the query position. We keep the writers, the duplicate-token, pooling, query, reader, brake and copier heads (67 in all) and the late MLPs, and replace the output of every other head at the query. Because the replacement happens only at the query, every other position, including the label tokens where the slots are built, runs as in the intact model. The test therefore asks whether the named heads are enough for the computation at the query, given intact slots. What it certifies is the readout from the slots; the evidence for how the slots are built is the handoffs and swaps of Section~\ref{sec:slot}.

\paragraph{The reference value.} The choice of reference matters a great deal. With means taken over natural text, the named set keeps nothing, and the margin goes to $-0.79$ / $-0.88$ nats from an intact $+1.13$ / $+0.55$ (rare / symbol labels). But a random set of the same size does no worse ($-0.83$ / $-0.79$), and neither does keeping no heads at all ($-0.82$ / $-0.81$). Replacing a thousand heads with a value 2.5 times further from their own output than the battery mean pushes the query residual away from anything the late layers have seen, and the test can no longer rank head sets. With the reference that \citet{wang2022ioi} use, the mean over a matched distribution, the test does rank them. ``Matched'' has to cover the label pair as well. The battery stores each prompt under two label pairs, and a reference that mixes them, or a resample partner that is the same prompt under the other pair, moves the results by up to a nat in either direction. Our first version of this control made exactly that mistake, and Appendix~\ref{app:reference} describes the error and how it biased the results. Every value below uses the mean over same-pair prompts at the query, or a resampled same-pair prompt.

\paragraph{Results.} With the named set plus the eight heads that a screen of layers 20--27 adds and the four early heads from the screen of layers 0--19 (79 heads; Appendix~\ref{app:complete}), the circuit alone keeps about 60\% of the margin on both label kinds. Under resampling the circuit-only margin sits $-0.28$ \ci{-0.50}{-0.09} below the intact $+0.55$ on symbol labels and $-0.47$ below $+1.14$ on rare ones, against $-0.65$ / $-0.98$ (symbol / rare) for a random set of about the same size drawn from the same layers and $-0.69$ / $-0.95$ for keeping no heads. Removing the complement one band at a time shows where the rest is: layers 0--19 cost $-0.16$ / $-0.29$, layers 20--27 cost $-0.07$ / $-0.10$, and layers 28--35 give $+0.09$ / $+0.17$. Removing the unnamed late heads helps, in other words, because on balance they act as brakes.

The natural-label study of Section~\ref{sec:transfer} later adds 110 heads, readers and brakes that matter when the label is a word. With that final set of 189 heads, the circuit alone does better than the intact model on both label kinds ($+0.89$ / $+1.26$ under resampling and $+0.93$ / $+1.62$ with the mean reference, symbol / rare; a random set gives $-0.64$ / $-0.98$). Removing the remaining heads of the two late bands helps ($+0.40$ / $+0.99$ and $+0.35$ / $+0.65$), and only the encoder band still costs anything ($-0.07$ / $-0.17$). Within the layers where the mechanism lives, then, the circuit is complete, and the same protocol closes it in SmolLM3-3B and Llama-3.2-3B as well (Appendix~\ref{app:complete}).

\paragraph{Completeness under selection.} A second sense of completeness asks whether the named sets carry their function on prompts they were not chosen on. Every named set was selected on one draw of prompts and tested on a fresh draw against random sets of the same size and layers. The readers cost $-0.74$ against $+0.02$, the brakes give $+1.39$ against $+0.22$, and the Llama sets give $+1.90$ and $-0.56$ against $+0.12$ and $0.00$.

\paragraph{Minimality.} Minimality is where our account departs from a sparse circuit. No single reader or brake is necessary in the intact network. The largest single-head cost in Qwen is 0.14 nats (the query head 23:26; the largest reader costs 0.09) and in Llama 0.09, and the largest single brake raises the margin by 0.13 and 0.12. \citet{wang2022ioi}'s criterion is weaker than this, since it lets a head count as necessary once the others are removed, and we have not run that conditional test. What Proposition~\ref{prop:gain} rules out is a single head carrying a 1.3-nat answer when no head writes more than 0.2 nats. In place of a minimal head set we report populations with direction-level tests, and we say so wherever this applies.

\subsection{Cutting the connections}\label{sec:arrows}
The interventions so far establish the parts, but the connections between them were inferred from the parts existing side by side. Four further experiments cut the connections directly, each on fresh prompt draws in Qwen3-4B (Table~\ref{tab:arrows}).

\paragraph{The readers pass their output to the late layers.} We take the concept battery and its class-flipped twin, and patch the twenty readers' outputs at the query from the twin into the intact run. This carries $-1.08$ nats of the twin's $-1.80$ on rare-word labels and $-0.54$ of $-0.95$ on symbol labels, so sixty percent of everything that differs between the two prompts passes through those twenty heads. We then hold the inputs of every attention head in layers 30--35 at their intact values, while the readers' outputs still come from the twin. The effect drops to $-0.81$ and $-0.35$. Holding the inputs of the layer 30--35 MLPs as well brings it down to $-0.39$ and $-0.19$. Of what the readers write, a quarter to a third is read by late attention heads (the copiers), about forty percent by the late MLPs, and the remaining third goes straight to the logits. The connection from readers to copiers is real, but it is one of three routes out of the readers.

\paragraph{The gate acts partly through the readers' queries.} We knock out the sixteen duplicate-token heads and add the gate direction $g$ back at layer 23 of the query. On unique-label prompts, where the gate is normally closed, this opens it. The margin rises by $+0.58$ nats at one unit of $g$ and by $+0.90$ at two, and the sharp reader \head{29}{11} moves its attention off the sink (0.91 to 0.37). We then remove the direction $g$ from the twenty readers' query inputs alone, leaving it everywhere else. The rise falls to $+0.43$ and $+0.61$, and \head{29}{11}'s sink mass climbs back (0.37 to 0.51; 0.35 to 0.62). Removing a random direction of the same kind from the same queries changes nothing ($+0.58$, $+0.92$, sink mass unchanged). A quarter to a third of what $g$ does to the answer therefore travels through the readers' queries, the route that our account draws, and the rest reaches the answer through heads and MLPs we have not named. On recurring-label prompts the same knockout lowers $g$ from $+18$ to $+13$ and costs $0.28$ nats, and adding one unit of $g$ brings the margin back to within $0.04$ of intact.

\paragraph{Amplifying a few readers does not replace the rest.} If the readers are many only because no single head can write enough (Proposition~\ref{prop:gain}), then amplifying a few of them should do the work of all. We multiplied the attention output of three readers (24:21, 29:11, 22:20) at the query by two, on prompts drawn with a seed not used to choose them. The margin rises by $+0.40$ (rare labels) and $+0.25$ (symbols), against $+0.04$ for three random heads. Multiplying by four still helps, and multiplying by eight breaks the model ($-0.56$, $-0.84$). Knocking out the other seventeen readers lowers the margin by $-0.89$ and $-0.50$, and amplifying the three by four on top of that recovers it only to $-0.46$ and $-0.44$; amplifying by eight makes it worse. Three amplified heads cannot stand in for seventeen. The gain bound holds and the readers' writes do add, but the other heads carry information that these three lack. Why training spread the computation this way remains open.

\paragraph{Position zero is a working sink.} Qwen and SmolLM3 add no beginning-of-sequence token, so position zero holds the prompt's first word. Zeroing its value in the reader layers costs $0.33$ nats on unique-label prompts and nothing measurable ($-0.03$) on recurring ones, and the same operation on position one costs nothing in either case. Zeroing it at every layer removes the sink altogether, since the readers then put no attention there, and the margin falls by $0.76$ and $0.43$. Position zero does the sink's job, and although its content is incidental to that job, the position is not empty. Appendix~\ref{app:arrows} refines this test and follows the part of $g$'s effect that does not pass through the readers.

\paragraph{The same cuts in SmolLM3 and Llama.} We repeated the two connection experiments on fresh draws in SmolLM3-3B and Llama-3.2-3B, using each model's own reader and duplicate-head sets and its own fitted gate direction (Table~\ref{tab:arrows_rep}). The readers' handoff replicates in both. SmolLM3's twenty readers carry the whole class-flipped twin effect ($-2.16$ against the twin's $-2.05$ on rare-word labels, $-1.52$ against $-1.30$ on symbols; sixteen random heads give $-0.11$ / $-0.07$), and Llama's twenty carry 45\% / 54\% of it ($-0.86$ of $-1.92$, $-0.46$ of $-0.86$). What happens after the readers differs by family. In SmolLM3, as in Qwen, holding the late attention inputs (L30--35) at their intact values removes 36\% of the handoff, and holding the late MLP inputs as well removes another 27\%, leaving 37\% that reaches the logits directly. In Llama the holds remove nothing ($-0.86 \to -1.02 \to -1.09$), so its readers reach the logits without a late relay.

The route from the gate to the readers' queries replicates in all three families. Adding $g$ at the query opens the gate in every family ($+0.53$ / $+0.69$ nats in SmolLM3 without and with the duplicate heads knocked out, $+0.64$ / $+0.92$ in Llama). Removing $g$ from the readers' queries takes back 12--46\% of that in SmolLM3 and 37--79\% in Llama, more at higher doses, while removing a random direction of the same norm takes back nothing ($+0.68$, $+0.91$). The step before the gate does not replicate as an effect on the margin. Knocking out the sixteen duplicate-token heads moves the gate projection in every family, in the direction the circuit predicts (on unique-label prompts by $+1.8$ units in SmolLM3 and $+4.5$ in Llama; on recurring ones by $-0.7$ and $-2.4$). In SmolLM3 and Llama, however, the margin does not fall with it ($+0.90$ and $-0.39$, where Qwen closes cleanly), because those heads also carry other information that the knockout removes. The connection from the duplicate heads into $g$ is therefore established at the level of the gate feature in all three models, and at the level of behaviour in Qwen only.

\begin{table}[t]
\centering\small
\caption{The connection cuts in SmolLM3-3B and Llama-3.2-3B, on fresh draws. Change in answer margin (nats) from intact, mean over prompts; concept battery rare / symbol ($n=71$ each), slotgate ($n=60$). The hold rows fix the inputs of the late attention and MLP layers (SmolLM3 L30--35, Llama L22--27) at their intact values.}\label{tab:arrows_rep}
\begin{tabular}{llrr}
\toprule
Intervention & Battery & SmolLM3 & Llama \\
\midrule
Class-flipped twin, everything & concept & $-2.05$ / $-1.30$ & $-1.92$ / $-0.86$ \\
\quad twenty readers' outputs from twin & concept & $-2.16$ / $-1.52$ & $-0.86$ / $-0.46$ \\
\quad\quad + hold late attention inputs & concept & $-1.38$ / $-1.24$ & $-1.02$ / $-0.61$ \\
\quad\quad + hold late attention and MLP inputs & concept & $-0.79$ / $-0.53$ & $-1.09$ / $-0.52$ \\
\quad random sixteen heads from twin & concept & $-0.11$ / $-0.07$ & $-0.16$ / $+0.05$ \\
\midrule
Twenty readers knocked / random twenty & slotgate & $-1.32$ / $-0.05$ & $-0.13$ / $+0.23$ \\
Duplicate heads knocked & slotgate & $+0.90$ & $-0.39$ \\
$g$ at the query ($\times 1$) & slotgate & $+0.53$ & $+0.64$ \\
\quad + duplicate heads knocked & slotgate & $+0.69$ & $+0.92$ \\
\quad\quad $g$ removed from readers' queries & slotgate & $+0.61$ & $+0.48$ \\
\quad\quad random direction removed & slotgate & $+0.68$ & $+0.91$ \\
\bottomrule
\end{tabular}
\end{table}

\begin{table}[t]
\centering\small
\caption{Four interventions on connections in Qwen3-4B-Base, on fresh draws. Change in answer margin (nats) from intact, mean over prompts. Concept battery: rare-word / symbol labels ($n=72$ each); slotgate: unique / recurring label prompts ($n=30$ each).}\label{tab:arrows}
\begin{tabular}{llr}
\toprule
Intervention & Battery & $\Delta$ margin \\
\midrule
Class-flipped twin, everything & concept & $-1.80$ / $-0.95$ \\
\quad twenty readers' outputs from twin & concept & $-1.08$ / $-0.54$ \\
\quad\quad + hold attention inputs L30--35 & concept & $-0.81$ / $-0.35$ \\
\quad\quad + hold attention and MLP inputs L30--35 & concept & $-0.39$ / $-0.19$ \\
\midrule
Duplicate heads knocked & slotgate & $+0.15$ / $-0.28$ \\
\quad + $g$ at layer 23 ($\times 1$) & slotgate & $+0.58$ / $+0.04$ \\
\quad + $g$, $g$ removed from readers' queries & slotgate & $+0.43$ / $-0.06$ \\
\quad + $2g$ & slotgate & $+0.90$ / $+0.12$ \\
\quad + $2g$, $g$ removed from readers' queries & slotgate & $+0.61$ / $-0.08$ \\
\quad + $2g$, random direction removed from readers' queries & slotgate & $+0.92$ / $+0.13$ \\
\midrule
Three readers $\times 2$ / random three $\times 2$ & concept & $+0.40$, $+0.04$ / $+0.25$, $+0.03$ \\
Other seventeen readers knocked & concept & $-0.89$ / $-0.50$ \\
\quad + three readers $\times 4$ & concept & $-0.46$ / $-0.44$ \\
\midrule
Position 0 value zeroed, L21--29 / position 1 & slotgate & $-0.33$, $-0.01$ / $-0.03$, $+0.02$ \\
Position 0 value zeroed, all layers & slotgate & $-0.76$ / $-0.43$ \\
\bottomrule
\end{tabular}
\end{table}

\section{One algorithm in three models}\label{sec:replication}
Three models trained by three groups on three corpora need not agree in every detail, so it helps to say in advance which details are free to vary. Some parts of the account follow from attention itself. A softmax over positions has to put its mass somewhere, so some position must serve as the sink, and a key written into a label token can only be read through that token's keys. These parts should appear in any transformer that does the task, and if a fourth model lacked them we would count the account as falsified. Other parts are left to training: which heads do each job, how many of them there are, how strong the gate is, and in which layers the work happens. These we expect to differ, and we report them as each model's implementation of the algorithm. We count a link as replicated when the same intervention has the same sign, with an interval that excludes zero, and the matched control stays within noise. We do not require the magnitudes to agree, and they differ by up to a factor of three.

By this standard the account replicates in SmolLM3-3B and Llama-3.2-3B (Table~\ref{tab:replication}). The value and the key of the slot dissociate in both (Section~\ref{sec:slot}). The read law fits 124 of 125 Llama slot-reading heads and predicts the one-slot regime more tightly than in Qwen, and the null action exists in both.

\begin{figure}[t]\centering
\fig{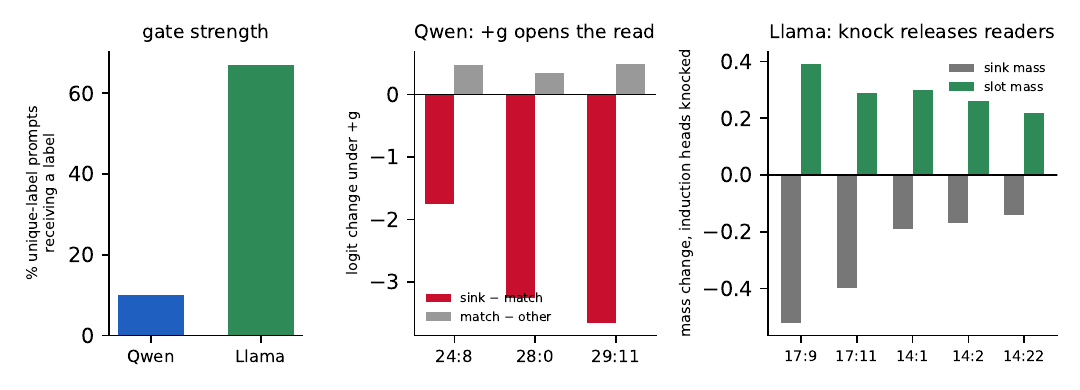}
\caption{The null action in two families. (a)~Llama-3.2-3B emits a label on 67\% of unique-label prompts and Qwen3-4B on 10\%: the same fallback, gated much more weakly in Llama. (b)~In Qwen, adding the gate signal $g$ at its natural dose lowers the sink logit relative to the matching slots at the three strongest readers (red), and moves matching relative to the other slots much less (grey). (c)~In Llama, knocking out the sixteen heads that a generic induction score selects moves the middle-layer readers' attention out of the sink and onto the slots, so in Llama the induction-scored heads hold the read closed.}
\label{fig:gatemodels}
\end{figure}

The implementations differ in ways worth recording. Llama's readers sit in layers 14--27 of 28, where Qwen's sit in 21--29 of 36. Llama's gate is weaker: 67\% of its unique-label prompts still receive a label, against 10\% in Qwen. In Llama the heads that a generic induction score selects act as a withholding population. They hold the middle-layer readers on the sink, and ablating them raises the margin by about a nat (Figure~\ref{fig:gatemodels}). And where Qwen has two nameable query heads, SmolLM3 and Llama spread the query computation over many.

The brakes show most clearly how a population can be invisible head by head and still exist as a set. Qwen's 24 brakes stand out in a single-head census, but Llama's do not. A census of the 288 Llama heads in layers 16--27 found no head whose knockout moves the margin by more than 0.12 nats, and 271 of them within $\pm 0.03$. Yet the sets exist. The 24 heads that the census ranks as brakes, knocked out together on a fresh twelve-demonstration draw, raise the margin by $+1.90$ \ci{+1.12}{+2.67}, against $+0.12$ for 24 random heads, and the 20 heads it ranks as readers cost $-0.56$ \ci{-1.01}{-0.12}, against $0.00$. At six demonstrations, where Llama's margin is only $+0.40$, the brake set gives $+0.57$ \ci{+0.19}{+0.93} against $+0.16$, and the reader set's effect is not detectable. Llama has the same two populations as Qwen, each weak head by head and strong as a set.

The algorithm also assembles in a fixed order during training. Across OLMo-2 checkpoints, copying a repeated label appears first, the null action second, and the readers' class selectivity last. The 1B model never develops the readers, which is why it never learns the task. This is the order the dependencies require, since nothing can be gated before there is something to copy, and nothing can be matched by class before the keys are written (Figure~\ref{fig:olmo}, Appendix~\ref{app:olmo}).

\begin{table}[t]\centering\small
\caption{Each link of the account in three models. \emph{Forced}: follows from the architecture and is expected in every model. \emph{Implementation}: set by training and reported per model. \emph{Named}: a small set of heads is necessary and sufficient. \emph{Population}: no single head is necessary, and the link is characterized by a direction with a size-matched random control. --: not run.}
\label{tab:replication}
\begin{tabularx}{\linewidth}{Yllll}
\toprule
Link & Kind & Qwen3-4B & SmolLM3-3B & Llama-3.2-3B \\
\midrule
Slot value (label) at label-token V & forced & yes & yes & yes \\
Slot key (class) at label-token K, mid layers & forced & yes & yes & yes \\
Writers, dose spread across depth & implementation & population & population & -- \\
Read law within regime / down-transfer & forced & 127/140, 0.07 & 7/8 & 124/125, 0.04 \\
Read law up-transfer (18 demos) & tested & failed & -- & failed \\
Duplicate detector feeds gate & implementation & named & yes & weak \\
Gate lowers null logit, not match & forced (sign) & yes & yes & -- \\
Query code & implementation & named & population & -- \\
Readers (source) & forced (population) & population & population & population \\
Brakes & implementation & 24 heads & suppressors & 24-head set, $+1.90$ \\
Value transport & implementation & 3 groups & -- & -- \\
Direct writer = late band & implementation & yes & yes & -- \\
Order term sign flip 6 vs 12 & implementation & yes & yes (3 carriers) & -- \\
\bottomrule
\end{tabularx}
\end{table}

\section{Natural concepts and meaningful labels}\label{sec:transfer}
The circuit was found on batteries whose classes are surface regularities and whose labels are arbitrary tokens. A mechanism worth the name should also carry retrieval when the class is a real semantic feature and when the labels are words with meanings of their own. To test this we built one more battery, \texttt{concept12n}. It has four concepts that a reader would call natural (topic: animal vs.\ food; language: English vs.\ Spanish; question vs.\ statement; implied sentiment: good vs.\ bad), twelve demonstrations each, and the same prompt written four ways that differ only in the two label words: arbitrary symbols (\texttt{foo}/\texttt{bar}), rare words (\texttt{zorp}/\texttt{quil}), the concept's own words (\texttt{animal}/\texttt{food}), and those words exchanged, which are the flipped labels of \citet{wei2023larger}. We ran 384 prompts per model, with the same named sets and interventions as in Section~\ref{sec:replication}, and wrote down our predictions before collecting the data (Appendix~\ref{app:claims}).

\paragraph{Arbitrary labels transfer.} With symbol and rare labels on the natural concepts, the named parts carry retrieval in all three families. In Qwen3-4B, keeping only the circuit (resampled reference) leaves the margin at intact ($-0.05$ and $+0.96$ nats relative to intact for symbol and rare labels). In Llama and SmolLM3 it falls below intact but stays far above a random set of the same size and layers: Llama gives $-0.58$ / $-0.11$ (random $-1.5$ / $-1.3$), and SmolLM3 gives $-0.82$ / $-0.47$ with intervals reaching zero (random $-1.6$ / $-1.5$). With the mean reference all three are at or above intact. With the final sets of this section (189, 122 and 163 heads), the symbol battery of Section~\ref{sec:accounting} gives the same answer in all three models: $+0.89$ / $+1.26$ in Qwen, $+0.87$ / $+1.01$ in SmolLM3 and $+0.03$ / $+0.75$ in Llama relative to intact under resampling (symbol / rare), against $-0.64$ / $-0.98$, $-0.71$ / $-1.11$ and $-0.26$ / $-0.71$ for random sets. The retrieval stage works the same way whatever the class is.

\paragraph{Word labels split the margin in two.} With the concept's own words as labels the intact margin triples, to $5.4$ nats in Qwen and SmolLM3 and $4.9$ in Llama. Because each prompt appears once with the natural labels and once with them exchanged, the two margins can be added and subtracted. Half their sum is the part of the margin that follows the demonstrations, which we call $R$, and half their difference is the part that follows what the words mean, which we call $P$. In all three models the two parts are about the same size ($R\approx2.7$--$2.9$, $P\approx2.2$--$2.7$ nats), which is why flipped labels end up near zero in a 3--4B base model, as \citet{wei2023larger} found behaviourally. The meaning part does not run through the circuit. Removing the MLPs outside the named late ones costs $3.8$ nats on natural labels against $1.1$ on symbols, and removing the late unnamed heads, which are brakes, \emph{helps} the flipped prompts ($+0.4$ nats in SmolLM3 and Llama), as we had predicted. The prior enters through the late band, and the circuit's readers have to win against it.

\paragraph{Word labels recruit more readers.} With only the named sets kept, a quarter to a half of $R$ survives on natural labels (the change in $R$ is $-2.16$ of $2.86$ nats in Qwen, $-1.50$ of $2.67$ in SmolLM3 and $-1.44$ of $2.71$ in Llama, with the reference matched on label pair and label set; see Appendix~\ref{app:reference}). Removing the complement band by band places the remainder in the reader and copier layers rather than somewhere new, and a head-by-head screen on the natural-label prompts identifies it. In Qwen, 73 heads outside the set in layers 16--27 cross $-0.03$ nats, and the largest of them (24:30, 27:18, 24:26, 26:4) put 0.4--0.7 of their attention on the slots whose label matches the answer. A second screen of layers 28--35, which the symbol-label screen had declared closed, finds 37 more. Two of these are slot-attending readers (33:30, 29:14), and the rest are sink-attending copiers and brakes. In Llama the screen finds 52 heads at layers 11--16, its writer and first-reader stage, and in SmolLM3 15 heads, nine of them in layers 14--17 and six in 30--31. On symbol prompts these heads attend in the same way but barely matter. The slot-reading stage is thus a population whose active membership depends on what has to be copied, since a word like \emph{animal} can be moved to the logits by many more heads than an arbitrary token can.

\begin{figure}[t]\centering
\includegraphics[width=\textwidth]{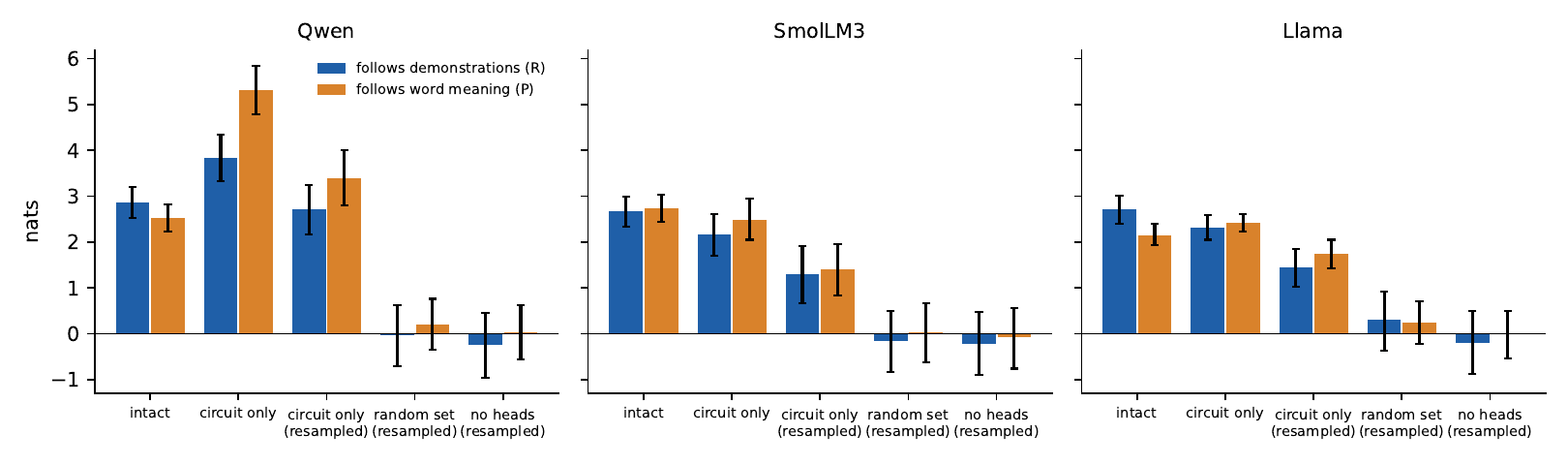}
\caption{Closing the circuit on natural labels. For each model, the part of the margin that follows the demonstrations ($R$, blue) and the part that follows what the label words mean ($P$, orange), intact and with only the named heads kept at the query (final sets of 189, 122 and 163 heads, with the late MLPs). The reference is the mean over same-kind prompts or a resampled same-kind prompt. The controls are a random set of the same size and layers, and no heads. Bars are means over 96 natural/flipped prompt pairs; whiskers are 95\% bootstrap intervals.}
\label{fig:closure}
\end{figure}

With the screened heads added (Qwen 189 heads, Llama 163, SmolLM3 122; Figure~\ref{fig:closure}), keeping only the circuit leaves the natural-label margin within the intact interval under resample ablation in Qwen ($-0.16$ nats of $2.87$, interval \ci{-0.71}{+0.38}; random set $-2.89$, no heads $-3.12$). In Llama and SmolLM3 it keeps 86\% and 81\% of $R$ under the mean reference ($-0.39$, $-0.51$; random $-2.4$, $-2.8$). Under the mean reference Qwen overshoots ($+0.97$), and its word-meaning part rises by $2.8$ nats. In all three models, removing the unnamed late heads raises $P$ by about $+2$ nats, so what remains outside the set at those layers is brakes, and the prior they hold down is the meaning of the word. Every layer band from the readers onward is closed in all three models (Qwen 28--35 $+1.13$, SmolLM3 30--35 $-0.01$, Llama 22--27 $-0.20$). What SmolLM3 and Llama still lose sits in the encoder band (SmolLM3 layers 0--17 $-0.56$, with only nine heads crossing the single-head threshold there), the same diffuse substrate we found with arbitrary labels.

Before the late screen we predicted that at most ten heads would cross the threshold and that the 28--35 complement would be neutral. Both predictions were wrong, and the misses are how we learned that the brake population grows with the meaning of the label. Resampled references are harsher than mean references on word labels, for reasons we give in Appendix~\ref{app:reference}.

\section{Model size and instruction tuning}\label{sec:scale}
Everything so far was measured in base models of three to four billion parameters. Two further questions have known behavioural answers but, until now, no mechanistic ones. \citet{wei2023larger} found that only large models follow flipped labels: a small model answers according to what the label word means, and a large one according to what the demonstrations say. And post-training is where every deployed model lives. We ran the same census in the Qwen3 family at 0.6B, 1.7B, 4B and 8B parameters, and in the Qwen3-4B instruct model using the base model's heads, again writing down our predictions first (Appendix~\ref{app:claims}). Each size got its own census, and neither head identities nor the gate direction were carried over from one size to another.

\begin{figure}[t]\centering
\includegraphics[width=\textwidth]{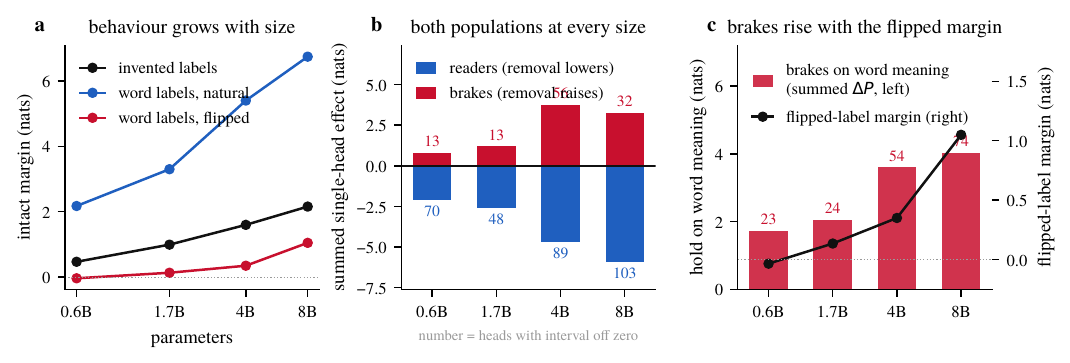}
\caption{One census at four sizes (Qwen3 base, 0.6B, 1.7B, 4B and 8B). (a)~Intact margins for invented labels, natural word labels and flipped word labels. (b)~Reader and brake populations: the summed effect of the heads whose single removal at the query lowers (blue) or raises (red) the margin with an interval that excludes zero; the number is the head count. (c)~The brakes' summed hold on word meaning (bars, left axis) against the flipped-label margin (line, right axis). The two rise together.}\label{fig:scale}
\end{figure}

\begin{figure}[t]\centering
\includegraphics[width=\textwidth]{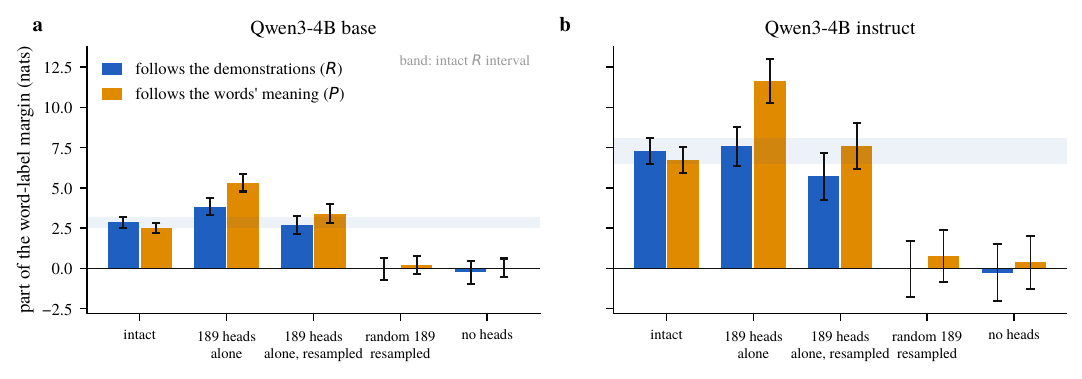}
\caption{Qwen3-4B base (a) and instruct (b) on the word-label prompts, with the base model's 189 heads in both. Blue is the part of the margin that follows the demonstrations ($R$), and orange the part that follows the words' meaning ($P$); the band is the intact $R$ interval. Post-training multiplies both parts by about 2.5 and keeps their ratio. The 189 heads alone, with a resampled reference, keep 79\% of $R$ in the instruct model, and a random set of the same size keeps none.}\label{fig:instruct}
\end{figure}

\paragraph{The same parts at every size.} The census screens every head in the upper half of the model (layers 10--27 of 28 at 0.6B and 1.7B, 16--35 of 36 at 4B and 8B) by removing it alone at the query on the \texttt{concept12n} battery, with the per-kind mean as reference. At every size there is a reader population, made up of heads whose removal lowers the margin with an interval that excludes zero, and a brake population of heads whose removal raises it. Table~\ref{tab:scale} and Figure~\ref{fig:scale} give the counts and summed effects. The readers' summed effect grows with size, from $-2.1$ nats at 0.6B to $-5.9$ at 8B. The brakes' summed effect grows from $+0.8$ through $+1.2$ to $+3.8$ at 4B and then falls back to $+3.3$ at 8B, where fewer heads cross the threshold (32 against 56). We had predicted a monotone rise, and that part of the prediction missed. The largest single reader, as a fraction of the intact margin, shrinks from 22\% to 18, 18 and 15\%. A bigger model spreads the read over more heads, as the gain bound of Section~\ref{sec:populations} predicts when the margin to be produced grows faster than any one head's output.

One thing we did not predict is that, within a layout, the parts sit at the same addresses. The two 28-layer models share their top brake, 22:7 ($+0.27$ and $+0.43$ nats), their second, 26:7, and nine reader positions. The two 36-layer models share their five strongest readers (29:11, 24:29, 28:0, 24:30, 24:22) and their five strongest brakes (25:30, 28:23, 31:30, 25:14, 33:21). From these runs we cannot tell whether this comes from initialisation, from data order or from something about the task.

\paragraph{The gate exists and acts at every size.} For each size we fitted the gate direction from that model's own pairs of unique- and recurring-label prompts (Section~\ref{sec:gate}). The direction separates held-out pairs perfectly at every layer of the upper half in all four models, and its size relative to the residual at the arrow peaks in the last third of the stack in all four (layer 22 of 28 at 0.6B and 1.7B, 23 of 36 at 4B, 24--34 of 36 at 8B). Adding the fitted direction at the query arrow of unique-label prompts, at the dose the model itself uses, raises their label margin by $+1.75$ (0.6B), $+0.95$ (1.7B) and $+1.11$ (8B) nats. At 4B, where the intact unique-label margin is already $+0.24$, the single dose gives $+0.25$ with an interval reaching zero, and the double dose gives $+0.78$. A random direction of the same norm moves the margin by at most $0.2$ in the wrong direction. Taking the direction away from recurring-label prompts lowers their margin by $1.35$, $0.69$, $0.72$ and $1.07$ nats at a single dose (0.6B, 1.7B, 4B, 8B), and by $2.9$, $1.7$, $1.8$ and $3.0$ at double. Two of our predictions failed here. We expected a single dose to clear zero at every size, and 4B did not. We also expected the double dose to beat the single one everywhere, but at 8B they tie ($1.10$ against $1.11$), because adding the direction saturates there while removing it still has twice the effect.

\paragraph{Where the flipped-label curve comes from.} At every size, the paired natural and flipped prompts give the part of the margin that follows the demonstrations ($R$) and the part that follows the words ($P$). Both grow with size, but $R$ grows faster. $P/R$ falls from $1.03$ at 0.6B through $0.92$ and $0.88$ to $0.73$ at 8B, and the flipped-label margin rises from $-0.03$ through $+0.14$ and $+0.35$ to $+1.05$ nats. This is the curve of \citet{wei2023larger}, reproduced inside a single family, and the census shows what moves it. The brakes on word meaning, the heads whose removal raises $P$, hold down $1.7$ nats at 0.6B, $2.1$ at 1.7B, $3.6$ at 4B and $4.0$ at 8B (23, 24, 54 and 74 heads). Of all the quantities in the census, this sum is the one that rises in step with the flipped-label margin at every size, and the reader population that carries $R$ grows alongside it. The gate does not track size in this way. Its dose effect is of the same order at every size (Table~\ref{tab:scale}), so the gate is a part present everywhere but not the lever that moves the curve.

The picture that emerges is a balance of strengths. In a small model the readers and the word-meaning prior are about equally strong, so exchanging the labels cancels the demonstrations. In a large model the readers out-pull the prior, and the brakes the model has grown hold the prior down further. Two details of our prediction missed: we expected the 1.7B flipped margin to be negative (it is $+0.14$, with an interval crossing zero) and the 0.6B rare-label margin to be below $0.3$ nats (it is $0.62$). The monotone parts of the prediction held.

\paragraph{Post-training keeps the circuit and turns up the brakes.} Qwen3-4B instruct has the same architecture as its base model, so the base model's 189-head set can be carried into it unchanged, with no re-identification. The set still carries retrieval. On the symbol battery the 189 heads alone, with a resampled reference, are at or above intact ($+0.43$ / $+1.29$ nats relative to intact for symbol and rare labels; base model $+0.89$ / $+1.26$), a random set of the same size and layers loses $0.66$ / $1.15$, and removing the 189 heads costs exactly what removing every head costs ($-0.61$ / $-1.00$ against $-0.60$ / $-1.00$). The base model's gate direction, without refitting, is positive at layer 23 on all 144 prompts.

What post-training changed is the gain (Figure~\ref{fig:instruct}). With word labels both parts of the margin are about $2.5$ times larger ($R=7.30$, $P=6.75$; base $2.87$ and $2.52$), and their ratio is unchanged ($0.92$ against $0.88$), so the instruct model still half-follows meaning on flipped labels, as \citet{wei2023larger} reported for instruction-tuned models. We had predicted that $P$ would rise and $R$ would stay flat, but both rose. The extra gain on meaning is held down by the unnamed heads of layers 16--27, whose removal now raises $P$ by $4.9$ nats (in the base model, about $2$), and the 189 heads keep 79\% of $R$ (in the base model, all of it). Post-training did not rewire the readers. It added braking in the population around them.

\begin{table}[t]\centering\small
\caption{One census at four sizes (Qwen3 base, \texttt{concept12n}, 384 prompts each, upper-half heads removed one at a time at the query). Readers: removal lowers the margin, with an interval that excludes zero; brakes: removal raises it. $R$ and $P$ are the parts of the word-label margin that follow the demonstrations and the meaning. Gate dose: change in the unique-label margin when the size's own fitted direction is added at the query arrow at the model's own dose and at twice that dose, and change in the recurring-label margin when it is taken away. 0.6B and 1.7B screen layers 10--27 (288 heads); 4B and 8B screen layers 16--35 (640 heads).}\label{tab:scale}
\begin{tabularx}{\linewidth}{Yrrrr}
\toprule
 & 0.6B & 1.7B & 4B & 8B \\
\midrule
Intact margin, symbol / rare labels (nats) & 0.32 / 0.62 & 0.87 / 1.12 & 1.62 / 1.57 & 2.43 / 1.88 \\
Natural / flipped word labels & 2.18 / $-0.03$ & 3.30 / $+0.14$ & 5.40 / $+0.35$ & 6.74 / $+1.05$ \\
$R$ / $P$ & 1.07 / 1.10 & 1.72 / 1.58 & 2.87 / 2.52 & 3.90 / 2.85 \\
Readers: count, summed effect & 70, $-2.08$ & 48, $-2.60$ & 89, $-4.69$ & 103, $-5.94$ \\
Largest reader, \% of intact & 22 & 18 & 18 & 15 \\
Brakes: count, summed effect & 13, $+0.82$ & 13, $+1.21$ & 56, $+3.76$ & 32, $+3.27$ \\
Brakes on word meaning: count, summed $\Delta P$ & 23, $+1.73$ & 24, $+2.06$ & 54, $+3.61$ & 74, $+4.03$ \\
Gate dose on unique-label margin, single / double & $+1.75$ / $+1.96$ & $+0.95$ / $+1.71$ & $+0.25$ / $+0.78$ & $+1.11$ / $+1.10$ \\
Gate removed from recurring-label prompts, single / double & $-1.35$ / $-2.87$ & $-0.69$ / $-1.68$ & $-0.72$ / $-1.80$ & $-1.07$ / $-3.02$ \\
\bottomrule
\end{tabularx}
\end{table}

\section{Building the circuit}\label{sec:build}
A mechanism found by taking a network apart should also be something one can build. We built this one twice: once by writing the weights of a small transformer by hand so that it implements exactly the circuit of Figure~\ref{fig:algorithm}, and once by training a small transformer on the task from scratch and looking for the parts.

\paragraph{Hand-set weights.} The hand-built model is attention-only, with three layers and a residual stream made of named blocks (class direction, label identity, position, token type, duplicate flag, gate, slot key and query code). Layer 1 holds the writers, which attend from each label token back to its sentence and copy the class into the slot-key block, and from the query arrow back to the query sentence into the query-code block. It also holds a duplicate detector that marks any label token whose identity appeared earlier. Layer 2 holds one pooling head, which averages the duplicate flags over the label positions into the gate block at the arrow. Layer 3 holds the readers. Their query is the query code plus the gate, their keys are the slot keys together with one fixed sink key at position 0, and their values are the label identities, which the unembedding reads out. Nothing is trained.

On the slotgate prompts the hand-built model reproduces Proposition~\ref{prop:law} exactly (largest deviation $10^{-16}$, since the law is the softmax it was built from), and it reproduces the behaviour of the gate described in Section~\ref{sec:gate}. On unique-label prompts the sink takes 0.67 of the readers' attention and the margin is $+0.24$, and adding $g$ empties the sink and raises the margin to $+0.73$. On recurring-label prompts (margin $+0.83$, sink 0.02), knocking out the duplicate detector sends 0.54 of the attention to the sink ($+0.39$), adding $g$ back rescues it ($+0.84$), and removing $g$ from the readers' queries undoes the rescue ($+0.39$). Knocking out the writers leaves nothing to read ($0.00$). Capping each reader's output as in Proposition~\ref{prop:gain} makes the margin grow with the number of readers (0.16, 0.49, 1.25 and 2.53 nats for 1, 3, 8 and 20 heads). This establishes sufficiency, in the sense that these parts, wired this way, produce the law and the gate. It says nothing yet about whether a trained model would choose them.

\paragraph{Trained from scratch.} We trained a four-layer, eight-head Llama-architecture model (hidden size 256, 3.4M parameters) on a synthetic version of the task. The vocabulary has 48 word tokens, each carrying two of eight latent classes. A sentence is three words that share exactly one class, 32 arbitrary label tokens are redrawn for every prompt, and each prompt has two to six demonstrations followed by a query. We used two training distributions with three seeds each: a mix of unique-label and recurring-label prompts, and recurring-label prompts only. The task is learned abruptly between 8k and 30k steps, and by 40k steps every run answers both kinds of prompt with a margin near 10 nats.

The parts we could look for with the same tools are present. Swapping a slot's residual stream at the writer layers moves the answer by 7--11 nats in every seed, and class-flip attribution finds one or two reader heads in the last layer. (A model this small needs no population, because its single heads clear the gain bound.) The gate, however, is absent under both distributions. The direction separating unique- from recurring-label prompts at the arrow has held-out sign accuracy of 0.45--0.62 in all six runs, steering along it does nothing consistent, and the two kinds of prompt are answered with the same margin (8.8--10.7 nats).

In hindsight our account predicts this. The small model has no prior to fall back on, and in its training data the unique-label mapping is always consistent, so retrieving is always right and nothing selects for a gate. We think the pretrained models have a gate because, against the language prior, a set of labels that never repeat does not certify a consistent mapping. The matching test is a third training distribution in which every unique-label prompt carries a fresh, unpredictable query label, so that recurrence is the only sign that retrieval will pay off. Four of five seeds learned the task, and in all four a gate appeared. Recurring-label prompts are answered with a margin of 8.7--9.1 nats and unique-label ones with 1.8--2.3, whereas under the other two distributions both kinds get 8.8--10.7. The recurring-minus-unique direction at the arrow in layer 2 is decodable on held-out prompts at 0.81--1.00 (0.45--0.62 in the controls), and steering along it by one natural gap moves the answer. Pushing a recurring prompt toward ``unique'' drops its margin from 9 to 0.7 and 1.1 in two seeds (a random direction of the same norm gives 7.0 and 5.8), and pushing a unique prompt toward ``recurring'' raises it from 2.3 to 8.0 in one seed and from 2.0 to 4.0 in another (random: 1.8 and 0.9). Which side can be steered varies by seed (Table~\ref{tab:toygate} in Appendix~\ref{app:toy}). In the mix and recurring-only controls the same push changes nothing, with margins moving by less than 0.1. A gate forms when, and only when, retrieval on unique-label prompts is penalised in training. This is the first causal evidence we know of for why the pretrained models have one.

\section{Related work}\label{sec:related}
\paragraph{What in-context learning computes.} One line of work asks what function few-shot ICL implements, abstracting away from the network. \citet{garg2022functions} train transformers to learn simple function classes in context, and \citet{akyurek2022linear} and \citet{vonoswald2022gd} show that, for linear regression, the computation can be read as gradient descent or ridge regression carried out by attention. These results describe the input--output map on synthetic tasks. We study a pretrained model on text and ask which heads carry out which step, and the answer we find is a lookup in a memory, with no step that resembles a gradient update.

\paragraph{Label tokens and the ICL circuit.} \citet{cho2024circuit} give a three-step outline of an ICL inference circuit in which label tokens store task information that a later head reads, and \citet{zheng2024labelwords} and \citet{hu2025addition} identify label words as the local carriers of that information. We add what the slot holds and on which side of attention, which heads write it and at what depth, and what switches the read on. \citet{kim2025schema} dissociate task schema from binding, whereas our dissociation is between a single label token's keys and values. \citet{jiao2026conflict} study what happens when demonstrations conflict, and the brakes we describe are the population that decides which of two recent labels wins.

\paragraph{Task vectors and induction heads.} \citet{hendel2023taskvectors} and \citet{todd2023fv} extract a task or function vector from a few heads' outputs at the query. Our query code is the same kind of object seen from the side of its consumers: two heads write it, and it lies in the span of the readers' match functionals and not along any one reader's weight-space probe. \citet{olsson2022induction} established induction heads as the copying mechanism of in-context learning and described the phase change in which they appear. Our duplicate-token detectors are the first half of that circuit put to a second use, producing the repeat signal that the gate pools. The readers are induction heads whose match is on a class feature, and the emergence order we observe extends the induction-head phase change by two further stages. \citet{yu2024qk} locate recency in query--key similarity to label positions; we find the same bias in the readers' attention, together with the population that cancels it.

\paragraph{Negative heads and the attention sink.} \citet{wang2022ioi} found negative name-mover heads that write against the answer, and \citet{mcdougall2023copysuppression} describe copy suppression as a general function of such heads. Our brakes are a population with that sign and a measurable job, which is to calibrate the model against recency. \citet{xiao2023sinks} describe the attention sink as a position that absorbs attention; we give it a function in retrieval, as the null position whose logit the gate controls.

\paragraph{Methods.} Activation patching descends from causal mediation analysis~\citep{vig2020mediation} and the locating experiments of \citet{meng2022rome}, and \citet{hanna2023greaterthan} and \citet{nanda2023grokking} use it to recover full circuits for arithmetic tasks. \citet{wang2022ioi} evaluate a circuit by faithfulness, completeness and minimality. Our exact accounting checks the decomposition prompt by prompt, our circuit-only runs test faithfulness, and our population results are where minimality fails, for a reason we can state. \citet{quirke2026stress} stress-test attribution with same-answer controls, which we use throughout. \citet{chen2024skillpath}, and the objection that ``circuit'' bundles together sparsity, localisability and modularity~\citep{bricken2026notcircuits}, frame the question that our split between components and populations answers (Appendix~\ref{app:populations}).

\section{Discussion}\label{sec:discussion}
\paragraph{Scope.} The account covers a narrow setting, and we want its boundary to be visible. It describes two-class label retrieval and function selection with short sentences and single-token labels, at up to twelve demonstrations, in three base models of three to four billion parameters. We re-found the parts and the gate at four sizes of one family (Section~\ref{sec:scale}), and the base model's heads still carry retrieval after post-training. The account does not cover instruction following, long-form tasks or binding among many classes, and our sizes stop at 8B. Within that boundary every link has an intervention and a control behind it. Outside it we have not looked.

\paragraph{Predictions that failed.} Some of our predictions did not hold, and the failures were among the more informative results. The read law does not extrapolate to longer prompts. Empirically, a head's null mass and its total matching mass stay roughly constant as matching slots are added, so the added slots share what was already there, and the law we fitted at twelve demonstrations had absorbed this saturation into its constants. We can describe this but cannot yet explain it. The fitted parameters are not separately identifiable ($n_{\mathrm{rest}}$ and $e^{L_{\mathrm{sink}}}$ enter only through their sum), so why each head's budgets stay fixed is a question that needs its own answer. We also expected that fine-tuning on many-shot prompts, where recency costs accuracy, would strengthen the brakes. Instead, 400 steps of LoRA tripled the readers' necessity and left the brakes where they were ($+1.32$ against $+1.44$ nats), so where the brakes come from in training is still open. A third test, knocking out the seven largest direct movers, had the predicted sign but an interval that includes zero.

\paragraph{Named parts and populations.} The most general lesson, we think, concerns the split between named components and populations. Where a step needs little output gain, the model concentrates it: two heads compute the query's class, sixteen detect repeats, and three value groups carry most of the label's identity. Where a step needs more gain than one head's $W_{OV}$ can provide, the model sums over many heads, and no single head is necessary. The populations are still circuits in the ordinary sense. Each has a definite input direction and a definite output direction, and those directions are specific when compared with random subspaces of the same dimension. What a population lacks is a single responsible head, and Proposition~\ref{prop:gain} explains why it cannot have one. We would suggest that an account of any transformer behaviour say, for each step, whether the model concentrated or summed, and give the population-level object where it summed. This is also our answer to the worry that ``circuit'' presupposes sparsity. In our account, how sparse each step is depends on the gain that step requires.

\paragraph{The null action.} A model that reads a memory in its prompt also needs a way not to read it, and the softmax supplies one for free, in the form of a position that every head can fall back on. In Qwen and SmolLM3 that position holds the prompt's first token, so being the sink is a learned role and the position is not empty. Removing its value at every layer leaves the readers nowhere to put the attention they do not spend, and unique-label prompts, where not reading is the right action, get worse by 0.4--0.8 nats. We suspect this is why attention sinks are found in so many models, and why the gate acts on the readers' queries and nowhere else. The query is the one place where a prompt-level signal can reach the only lever there is.

\paragraph{Implications.} Some consequences reach beyond the setting we studied, though they rest on one family of tasks and should be read with that in mind.

In this setting, in-context learning is retrieval. We found no step that resembles fitting a rule or taking a gradient step. The model binds each label to the class of the sentence before it and looks that class up when the query arrives. What a model can learn from examples is then limited by what its encoder already represents. In our batteries, a class that was not linearly readable at the arrow (parity, the third letter of a word, digit rules) never became a usable key, however many demonstrations the prompt held, and writing the key in by hand raised accuracy on such prompts from 50 to 71\%. The link is correlational, but it suggests a cheap diagnostic for few-shot tasks: probe for the class at the arrow before writing the prompt.

The behaviour that \citet{wei2023larger} describe as appearing with scale appears here without any new component. Readers, brakes and a working gate are all present at 0.6B, and what changes with size is the balance between the population that follows the demonstrations and the population that holds down what the label words mean. If other abilities that seem to switch on with scale work the same way, they would be better understood by measuring how the strengths of existing parts shift with size than by searching for a part that is missing in small models. Post-training, at least in the one model we examined, also worked on gain. It left the base model's circuit in place and changed how strongly its parts pull, which is encouraging for anyone hoping that mechanisms found in base models will carry over to the models people actually use.

Finally, the model carries a low-dimensional signal that says whether its prompt contains a mapping worth following, and we can add that signal or take it away. Whether a model trusts its context or its training is a question that comes up in retrieval-augmented generation, in prompts that try to override a model with demonstrations, and whenever a user supplies information that contradicts what the model learned. Our results suggest that, at least for labelled demonstrations, part of this decision is made by a few identifiable directions and populations, which could be measured and perhaps adjusted directly.

It has not escaped our notice that the gate we have described, a signal that tells the model whether its prompt holds a memory worth reading, suggests a possible mechanism by which a language model decides whether to consult its context at all, before other heads weigh what the context says against what the model already knows.

\section*{Acknowledgements}
A.G. is grateful to the organizers, instructors and supporters of the Atlas Fellowship, where his interest in mechanistic interpretability began.

\section*{Reproducibility}
Code, batteries, the per-prompt measurement hooks and the scripts that produce every figure and table are released at \url{https://github.com/andrewgcodes/in-context-learning-circuit}. All runs used single H100s with the HuggingFace implementations of the models in bfloat16.

\raggedbottom
\bibliographystyle{iclr2024_conference}
\bibliography{refs}

\begin{thebibliography}{25}
\providecommand{\natexlab}[1]{#1}
\providecommand{\url}[1]{\texttt{#1}}
\expandafter\ifx\csname urlstyle\endcsname\relax
  \providecommand{\doi}[1]{doi: #1}\else
  \providecommand{\doi}{doi: \begingroup \urlstyle{rm}\Url}\fi

\bibitem[Akyürek et~al.(2022)Akyürek, Schuurmans, Andreas, Ma, and
  Zhou]{akyurek2022linear}
Ekin Akyürek, Dale Schuurmans, Jacob Andreas, Tengyu Ma, and Denny Zhou.
\newblock What learning algorithm is in-context learning? investigations with
  linear models.
\newblock \emph{arXiv preprint arXiv:2211.15661}, 2022.

\bibitem[Brown et~al.(2020)Brown, Mann, Ryder, Subbiah, Kaplan, Dhariwal,
  Neelakantan, Shyam, Sastry, Askell, Agarwal, Herbert-Voss, Krueger, Henighan,
  Child, Ramesh, Ziegler, Wu, Winter, Hesse, Chen, Sigler, Litwin, Gray, Chess,
  Clark, Berner, McCandlish, Radford, Sutskever, and Amodei]{brown2020fewshot}
Tom~B. Brown, Benjamin Mann, Nick Ryder, Melanie Subbiah, Jared Kaplan,
  Prafulla Dhariwal, Arvind Neelakantan, Pranav Shyam, Girish Sastry, Amanda
  Askell, Sandhini Agarwal, Ariel Herbert-Voss, Gretchen Krueger, Tom Henighan,
  Rewon Child, Aditya Ramesh, Daniel~M. Ziegler, Jeffrey Wu, Clemens Winter,
  Christopher Hesse, Mark Chen, Eric Sigler, Mateusz Litwin, Scott Gray,
  Benjamin Chess, Jack Clark, Christopher Berner, Sam McCandlish, Alec Radford,
  Ilya Sutskever, and Dario Amodei.
\newblock Language models are few-shot learners.
\newblock \emph{arXiv preprint arXiv:2005.14165}, 2020.

\bibitem[Chen et~al.(2024)Chen, Zhu, Yang, and Wang]{chen2024skillpath}
Hang Chen, Jiaying Zhu, Xinyu Yang, and Wenya Wang.
\newblock Skill path: Unveiling language skills from circuit graphs.
\newblock \emph{arXiv preprint arXiv:2410.01334}, 2024.

\bibitem[Cho et~al.(2024)Cho, Kato, Sakai, and Inoue]{cho2024circuit}
Hakaze Cho, Mariko Kato, Yoshihiro Sakai, and Naoya Inoue.
\newblock Revisiting in-context learning inference circuit in large language
  models.
\newblock \emph{arXiv preprint arXiv:2410.04468}, 2024.

\bibitem[Elhage et~al.(2021)Elhage, Nanda, Olsson, Henighan, Joseph, Mann,
  Askell, Bai, Chen, Conerly, et~al.]{elhage2021framework}
Nelson Elhage, Neel Nanda, Catherine Olsson, Tom Henighan, Nicholas Joseph, Ben
  Mann, Amanda Askell, Yuntao Bai, Anna Chen, Tom Conerly, et~al.
\newblock A mathematical framework for transformer circuits.
\newblock Transformer Circuits Thread, 2021.
\newblock URL \url{https://transformer-circuits.pub/2021/framework/index.html}.

\bibitem[Garg et~al.(2022)Garg, Tsipras, Liang, and Valiant]{garg2022functions}
Shivam Garg, Dimitris Tsipras, Percy Liang, and Gregory Valiant.
\newblock What can transformers learn in-context? a case study of simple
  function classes.
\newblock \emph{arXiv preprint arXiv:2208.01066}, 2022.

\bibitem[Hanna et~al.(2023)Hanna, Liu, and Variengien]{hanna2023greaterthan}
Michael Hanna, Ollie Liu, and Alexandre Variengien.
\newblock How does gpt-2 compute greater-than?: Interpreting mathematical
  abilities in a pre-trained language model.
\newblock \emph{arXiv preprint arXiv:2305.00586}, 2023.

\bibitem[Hendel et~al.(2023)Hendel, Geva, and Globerson]{hendel2023taskvectors}
Roee Hendel, Mor Geva, and Amir Globerson.
\newblock In-context learning creates task vectors.
\newblock \emph{arXiv preprint arXiv:2310.15916}, 2023.

\bibitem[Hu et~al.(2025)Hu, Yin, Jordan, Steinhardt, and Chen]{hu2025addition}
Xinyan Hu, Kayo Yin, Michael~I. Jordan, Jacob Steinhardt, and Lijie Chen.
\newblock Understanding in-context learning of addition via activation
  subspaces.
\newblock \emph{arXiv preprint arXiv:2505.05145}, 2025.

\bibitem[Jiao et~al.(2026)Jiao, Wang, and Hu]{jiao2026conflict}
Difan Jiao, Di~Wang, and Lijie Hu.
\newblock Understanding the dynamics of demonstration conflict in in-context
  learning.
\newblock \emph{arXiv preprint arXiv:2603.04464}, 2026.

\bibitem[Kim(2025)]{kim2025schema}
Chaeha Kim.
\newblock Task schema and binding: A double dissociation study of in-context
  learning.
\newblock \emph{arXiv preprint arXiv:2512.17325}, 2025.

\bibitem[{LessWrong post}(2026)]{bricken2026notcircuits}
{LessWrong post}.
\newblock What if not circuits?
\newblock LessWrong, 2026.
\newblock URL
  \url{https://www.lesswrong.com/posts/mMERyrvEJ4xbiozie/what-if-not-circuits}.

\bibitem[McDougall et~al.(2023)McDougall, Conmy, Rushing, McGrath, and
  Nanda]{mcdougall2023copysuppression}
Callum McDougall, Arthur Conmy, Cody Rushing, Thomas McGrath, and Neel Nanda.
\newblock Copy suppression: Comprehensively understanding an attention head.
\newblock \emph{arXiv preprint arXiv:2310.04625}, 2023.

\bibitem[Meng et~al.(2022)Meng, Bau, Andonian, and Belinkov]{meng2022rome}
Kevin Meng, David Bau, Alex Andonian, and Yonatan Belinkov.
\newblock Locating and editing factual associations in gpt.
\newblock \emph{arXiv preprint arXiv:2202.05262}, 2022.

\bibitem[Nanda et~al.(2023)Nanda, Chan, Lieberum, Smith, and
  Steinhardt]{nanda2023grokking}
Neel Nanda, Lawrence Chan, Tom Lieberum, Jess Smith, and Jacob Steinhardt.
\newblock Progress measures for grokking via mechanistic interpretability.
\newblock \emph{arXiv preprint arXiv:2301.05217}, 2023.

\bibitem[Olsson et~al.(2022)Olsson, Elhage, Nanda, Joseph, DasSarma, Henighan,
  Mann, Askell, Bai, Chen, Conerly, Drain, Ganguli, Hatfield-Dodds, Hernandez,
  Johnston, Jones, Kernion, Lovitt, Ndousse, Amodei, Brown, Clark, Kaplan,
  McCandlish, and Olah]{olsson2022induction}
Catherine Olsson, Nelson Elhage, Neel Nanda, Nicholas Joseph, Nova DasSarma,
  Tom Henighan, Ben Mann, Amanda Askell, Yuntao Bai, Anna Chen, Tom Conerly,
  Dawn Drain, Deep Ganguli, Zac Hatfield-Dodds, Danny Hernandez, Scott
  Johnston, Andy Jones, Jackson Kernion, Liane Lovitt, Kamal Ndousse, Dario
  Amodei, Tom Brown, Jack Clark, Jared Kaplan, Sam McCandlish, and Chris Olah.
\newblock In-context learning and induction heads.
\newblock \emph{arXiv preprint arXiv:2209.11895}, 2022.

\bibitem[Quirke(2026)]{quirke2026stress}
Philip Quirke.
\newblock Necessary, decodable and reversible, yet not transferable: A stress
  test for attention-head role claims.
\newblock \emph{arXiv preprint arXiv:2606.08292}, 2026.

\bibitem[Todd et~al.(2023)Todd, Li, Sharma, Mueller, Wallace, and
  Bau]{todd2023fv}
Eric Todd, Millicent~L. Li, Arnab~Sen Sharma, Aaron Mueller, Byron~C. Wallace,
  and David Bau.
\newblock Function vectors in large language models.
\newblock \emph{arXiv preprint arXiv:2310.15213}, 2023.

\bibitem[Vig et~al.(2020)Vig, Gehrmann, Belinkov, Qian, Nevo, Sakenis, Huang,
  Singer, and Shieber]{vig2020mediation}
Jesse Vig, Sebastian Gehrmann, Yonatan Belinkov, Sharon Qian, Daniel Nevo,
  Simas Sakenis, Jason Huang, Yaron Singer, and Stuart Shieber.
\newblock Causal mediation analysis for interpreting neural nlp: The case of
  gender bias.
\newblock \emph{arXiv preprint arXiv:2004.12265}, 2020.

\bibitem[von Oswald et~al.(2022)von Oswald, Niklasson, Randazzo, Sacramento,
  Mordvintsev, Zhmoginov, and Vladymyrov]{vonoswald2022gd}
Johannes von Oswald, Eyvind Niklasson, Ettore Randazzo, João Sacramento,
  Alexander Mordvintsev, Andrey Zhmoginov, and Max Vladymyrov.
\newblock Transformers learn in-context by gradient descent.
\newblock \emph{arXiv preprint arXiv:2212.07677}, 2022.

\bibitem[Wang et~al.(2022)Wang, Variengien, Conmy, Shlegeris, and
  Steinhardt]{wang2022ioi}
Kevin Wang, Alexandre Variengien, Arthur Conmy, Buck Shlegeris, and Jacob
  Steinhardt.
\newblock Interpretability in the wild: a circuit for indirect object
  identification in gpt-2 small.
\newblock \emph{arXiv preprint arXiv:2211.00593}, 2022.

\bibitem[Wei et~al.(2023)Wei, Wei, Tay, Tran, Webson, Lu, Chen, Liu, Huang,
  Zhou, and Ma]{wei2023larger}
Jerry Wei, Jason Wei, Yi~Tay, Dustin Tran, Albert Webson, Yifeng Lu, Xinyun
  Chen, Hanxiao Liu, Da~Huang, Denny Zhou, and Tengyu Ma.
\newblock Larger language models do in-context learning differently.
\newblock \emph{arXiv preprint arXiv:2303.03846}, 2023.

\bibitem[Xiao et~al.(2023)Xiao, Tian, Chen, Han, and Lewis]{xiao2023sinks}
Guangxuan Xiao, Yuandong Tian, Beidi Chen, Song Han, and Mike Lewis.
\newblock Efficient streaming language models with attention sinks.
\newblock \emph{arXiv preprint arXiv:2309.17453}, 2023.

\bibitem[Yu \& Ananiadou(2024)Yu and Ananiadou]{yu2024qk}
Zeping Yu and Sophia Ananiadou.
\newblock How do large language models learn in-context? query and key matrices
  of in-context heads are two towers for metric learning.
\newblock \emph{arXiv preprint arXiv:2402.02872}, 2024.

\bibitem[Zheng et~al.(2024)Zheng, Ma, Lin, and Yang]{zheng2024labelwords}
Bowen Zheng, Ming Ma, Zhongqiao Lin, and Tianming Yang.
\newblock Label words as local task vectors in in-context learning.
\newblock \emph{arXiv preprint arXiv:2406.16007}, 2024.

\end{thebibliography}

\appendix
\section{Batteries, interventions and statistics}\label{app:setup}
\paragraph{Batteries.} Full sentence pools, label sets and seeds are in the released code. The concept battery draws three demonstrations per class from two pools of a given axis, and concept12 draws six per class. Slotgate draws two per class and pairs each prompt with a token-identical unique-label twin. Manyclass draws $K$ classes from six pools (positive, negative, animals, food, Spanish, questions), with $12/K$ (or $18/K$) demonstrations per class and arbitrary single-token labels.
\paragraph{Statistics.} All intervals are paired bootstraps over prompts (4000 resamples). Head sets selected on a battery were confirmed on a fresh draw of prompts from the same pools with a different seed. Random controls match the number of heads and their layers.
\paragraph{Tokenization.} Targets are scored as full token sequences, and the first token of a two-token label carries the key. Llama's tokenizer prepends a beginning-of-sequence token, which the prompt-side comparison strips.

\section{Table of effects}\label{app:claims}
Every number in the main text, with its model, $n$ (prompts) and paired-bootstrap interval.
{\scriptsize
\begin{xltabular}{\linewidth}{>{\hsize=.72\hsize}Y l r >{\hsize=1.28\hsize}Y}
\caption{Effects behind the main text (nats; paired against the intact run; 95\% bootstrap intervals).}\label{tab:claims}\\

\toprule
Claim & Model & $n$ & Effect [95\% CI] \\
\midrule
\endfirsthead
\multicolumn{4}{l}{\textit{Table \ref{tab:claims} continued}}\\

\toprule
Claim & Model & $n$ & Effect [95\% CI] \\
\midrule
\endhead
\midrule\multicolumn{4}{r}{\textit{continued on next page}}\\
\endfoot
\endlastfoot
Label exchange: K/V at all positions, reader layers & Qwen & 144 & $-7.15$ [$-7.39,-6.90$] \\
Label exchange: label-token K/V, all layers (= full twin) & Qwen & 144 & $-7.18$ [$-7.43,-6.94$] \\
Label exchange: 8 readers' V / K & Qwen & 144 & $-2.16$ [$-2.41,-1.90$] / $+0.56$ [$+0.44,+0.69$] \\
Label exchange: label-token K/V, all layers (= full twin) & Llama & 144 & $-4.92$ [$-5.31,-4.55$] \\
Label exchange: label-token K/V, all layers (= full twin) / L21--29 & SmolLM3 & 144 & $-5.55$ [$-5.93,-5.15$] / $-2.94$ [$-3.30,-2.58$] \\
Sentence swap: twin & Qwen & 176 & $-2.87$ [$-3.13,-2.62$] \\
Sentence swap: label K/V L21--29 & Qwen & 176 & $-2.35$ [$-2.58,-2.12$] \\
Sentence swap: 8 readers' K / V & Qwen & 176 & $-0.55$ [$-0.67,-0.44$] / $+0.10$ [$+0.07,+0.14$] \\
Sentence swap: label K/V by band & Llama & 432 & $-0.02$, $-0.15$, $-0.15$, $-0.04$ (twin $-0.23$) \\
Sentence swap: query attention by band & Llama & 432 & $-0.03$, $-0.18$, $-0.13$ \\
Sentence swap: exact span & SmolLM3 & 432 & $-0.33$ [$-0.41,-0.25$] \\
Same-class twin (control) & Qwen & 168 & $+0.08$ [$-0.07,+0.23$] / $+0.08$ \\
Spread (4 layers) vs single-layer dose, natural / double & SmolLM3 & 144 & $+0.56$ [$+0.48,+0.65$] vs $+0.37$ [$+0.29,+0.46$] / $+0.69$ [$+0.58,+0.81$] vs $+0.20$ [$+0.09,+0.31$] \\
Gate signal, recurring / unique & Qwen & 30/30 & $+18.3$ / $-20.9$ \\
Gate signal, 16 duplicate heads off & Qwen & 30/30 & $-0.3$ / $-19.1$ (random16 $+17.5$ / $-21.3$; induction16 $+14.0$ / $-7.4$) \\
$+g$: null $-$ match logit & Qwen & 30 & $-1.75$ [$-1.94,-1.55$] / $-3.25$ / $-3.65$; random $+0.17$ / $-0.17$ / $-0.19$ \\
$+g$: match $-$ other logit & Qwen & 30 & $+0.47$ / $+0.34$ / $+0.48$; random $-0.04$ / $+0.05$ / $+0.03$ \\
Query code: pair handoff & Qwen & 144 & $-0.53$ [$-0.68,-0.39$] (exact twin $-1.33$) \\
Query code: hold attn 24--29 / 24--35 & Qwen & 144 & $-0.14$ [$-0.17,-0.10$] / $-0.04$ [$-0.06,-0.03$] \\
Query code: hold attn+MLP / MLP 30--35 & Qwen & 144 & $-0.01$ / $-0.66$ [$-0.83,-0.50$] \\
Readers 21--29 attention out & Qwen & 144 & $-1.10$ [$-1.39,-0.82$] \\
Reader-20 knock, fresh draw & Qwen & 144 & $-0.74$ [$-0.95,-0.55$] / $+0.02$ \\
Brake-24 knock, fresh draw & Qwen & 144 & $+1.39$ [$+0.88,+1.88$] / $+0.22$ \\
Top-7 direct movers knock & Qwen & 144 & $-0.16$ [$-0.56,+0.24$] / $-0.02$, $+0.13$, $+0.05$ \\
Brake-24 after many-shot / few-shot LoRA & Qwen & 144 & $+1.32$ [$+0.81,+1.82$] / $+1.44$ [$+0.86,+2.01$] \\
Read law within regime & Qwen, Llama & 96 & 127/140, 124/125 heads within 0.1 \\
Read law, one slot per class & Qwen, Llama & 96 & median error 0.07 / 0.04 \\
Read law, 18 demonstrations & Qwen, Llama & 96 & median error 0.15 / 0.13 (failed) \\
Exact accounting & Qwen, SmolLM3 & 144 & recon.\ error 0.045 / 0.048; 14/46 and 8/45 heads \\
Llama single-head census & Llama & 144 & max reader $-0.09$, max brake $+0.12$, random-288 joint $+0.74$ \\
Census-ranked brake-24 / reader-20 joint knock, 12 demos & Llama & 144 & $+1.90$ [$+1.12,+2.67$] (random $+0.12$) / $-0.56$ [$-1.01,-0.12$] (random $0.00$) \\
Census-ranked brake-24 / reader-20 joint knock, 6 demos & Llama & 432 & $+0.57$ [$+0.19,+0.93$] (random $+0.16$) / $-0.06$ [$-0.32,+0.21$] \\
\midrule
Instruct, 189 base heads only (resampled) vs intact, symbol / rare & Qwen3-4B instruct & 144 & $+0.43$ [$-0.17,+1.01$] / $+1.29$ [$-0.08,+2.69$] (random $-0.66$ / $-1.15$) \\
Instruct, 189 heads removed / all heads removed & Qwen3-4B instruct & 144 & $-0.61$ / $-1.00$ vs $-0.60$ / $-1.00$ \\
Instruct, word labels: $R$ / $P$ intact & Qwen3-4B instruct & 96 & $+7.30$ [$+6.49,+8.09$] / $+6.75$ [$+5.92,+7.55$] \\
Instruct, 189 heads only (resampled): $\Delta R$ / $\Delta P$ & Qwen3-4B instruct & 96 & $-1.53$ [$-3.07,-0.11$] / $+0.88$ [$-0.57,+2.29$] (random $-7.29$ / $-5.98$) \\
Instruct, unnamed heads L16--27 removed: $\Delta R$ / $\Delta P$ & Qwen3-4B instruct & 96 & $-0.27$ / $+4.93$ \\
Gate dose $+g$ at query, unique-label prompts, single / double & Qwen3-0.6B & 30 & $+1.75$ [$+1.57,+1.95$] / $+1.96$ [$+1.68,+2.24$]; random $-0.13$ / $-0.26$ \\
 & Qwen3-1.7B & 30 & $+0.95$ [$+0.68,+1.18$] / $+1.71$ [$+1.33,+2.08$]; random $+0.02$ / $-0.07$ \\
 & Qwen3-4B & 30 & $+0.25$ [$-0.14,+0.62$] / $+0.78$ [$+0.32,+1.25$]; random $-0.08$ / $-0.34$ \\
 & Qwen3-8B & 30 & $+1.11$ [$+0.79,+1.43$] / $+1.10$ [$+0.57,+1.63$]; random $+0.09$ / $+0.20$ \\
Gate removed at query, recurring-label prompts, single / double & Qwen3 0.6B / 1.7B / 4B / 8B & 30 & $-1.35$ / $-2.87$; $-0.69$ / $-1.68$; $-0.72$ / $-1.80$; $-1.07$ / $-3.02$ \\
Census, intact symbol / rare / natural / flipped & Qwen3 0.6B / 1.7B / 4B / 8B & 384 & 0.32 / 0.62 / 2.18 / $-0.03$; 0.87 / 1.12 / 3.30 / $+0.14$; 1.62 / 1.57 / 5.40 / $+0.35$; 2.43 / 1.88 / 6.74 / $+1.05$ \\
Census, readers: count, summed effect & Qwen3 0.6B / 1.7B / 4B / 8B & 384 & 70, $-2.08$; 48, $-2.60$; 89, $-4.69$; 103, $-5.94$ \\
Census, brakes: count, summed effect & Qwen3 0.6B / 1.7B / 4B / 8B & 384 & 13, $+0.82$; 13, $+1.21$; 56, $+3.76$; 32, $+3.27$ \\
Census, brakes on word meaning: count, summed $\Delta P$ & Qwen3 0.6B / 1.7B / 4B / 8B & 96 pairs & 23, $+1.73$; 24, $+2.06$; 54, $+3.61$; 74, $+4.03$ \\
Gate fit, held-out sign agreement & Qwen3 0.6B / 1.7B / 8B & 23 pairs & 1.00 at every layer 6--27 / 6--27 / 10--35 \\
\bottomrule
\end{xltabular}}

\section{The encoder bound and assembly from parts}\label{app:encoder}
The encoder bound is correlational. Every axis that failed to bind (parity, third letter, letter set, digit rules) is one that the model never represents linearly at the arrow, at any layer. Language, by contrast, becomes linear by layer 8 and binds weakly through layers 10--20, and topic, sentiment and implied meaning become linear by layer 13 and bind through layers 21--29. Adding the gate to a prompt that does not bind changes nothing.

For assembly from parts, we wrote the task direction $u$ into the query-class slots at layer 16 and at the query of a parity prompt. This raises accuracy from 50\% to 71\% (question 38\%$\to$58\%, vowel 50\%$\to$71\%), while key-only, query-only and random directions stay within $\pm0.1$, and writing into the wrong-class slots hurts. SmolLM3 reproduces this at a model-specific dose. For the same-answer control, handoffs from a same-class twin of the same length move the answer by at most 0.06 nats.

\section{Why the all-layer label-token handoff must reproduce the swap}\label{app:induction}
Let two prompts $A$ and $B$ have the same length and identical tokens except at a set of positions $S$ (here the label tokens). Run the model on $B$ and, at every layer, replace the keys and values computed at the positions in $S$ by those from the run on $A$. We claim that every position outside $S$ then carries exactly its $A$-run residual stream at every layer, so the logits at the query equal those of prompt $A$.

The proof is an induction over layers. At the input, positions outside $S$ have the same token in $A$ and $B$, so their embeddings agree. Suppose that at the input of layer $\ell$ every position outside $S$ agrees with the $A$ run. Attention at such a position reads its own query, which agrees, and keys and values from every position. Those outside $S$ agree by the hypothesis, and those inside $S$ have been replaced by the $A$ run's. So the attention output agrees, the MLP, which acts position by position, agrees, and the residual stream at the output of layer $\ell$ agrees. The query position is outside $S$, so its final residual and its logits are those of $A$. Nothing in the argument uses what the model has learned. It holds for a randomly initialised transformer and for any choice of $S$, including a set of positions that carries no task information at all. A reviewer pointed this out to us and checked it on random transformers with and without query--key normalisation.

The exact recovery in Section~\ref{sec:slot} therefore tells us only that the harness patches what it says it patches. The informative handoffs are the ones the argument does not cover: keys and values at the label tokens in a band of layers, values without keys, keys without values, and a changed sentence transmitted through the keys at positions where the input did not change.

\section{Order effects and the many-shot fine-tuning test}\label{app:order}
A model that has stored its demonstrations in a memory should not care about their order, and on average it does not: permuting the twelve demonstrations leaves the mean margin unchanged. Single prompts do move, by more than a nat on half of them, and the readers are the source of the movement. Their attention favours late slots, with a gradient of 0.5--0.85 nats of logit from the first slot to the last, so a prompt whose last matching demonstration sits near the query is read more strongly than one whose matching demonstrations come early. The brakes cancel this. With the 24 brakes knocked out the model prefers whichever label came last. At six demonstrations the brakes carry the anti-recency term outright, and at twelve they hold the model order-neutral against the readers' pull. In SmolLM3 three heads carry most of the order term and can be named.

We predicted that fine-tuning on many-shot prompts, where recency costs accuracy, would strengthen the brakes. It did not. We trained two LoRA adapters on Qwen3-4B (rank 16, $\alpha = 32$, learning rate $10^{-4}$, 400 steps, adapters on every linear layer) on axes disjoint from the test battery, one on many-shot prompts and one on few-shot prompts, so that the two arms differ only in the number of demonstrations. After training, knocking out the 24 brakes raises the margin by $+1.32$ \ci{+0.81}{+1.82} in the many-shot arm and $+1.44$ \ci{+0.86}{+2.01} in the few-shot arm, so the brakes and their gain are unchanged. The readers moved instead: the cost of knocking them out rose about threefold under adaptation in both arms. Whatever sets the brakes' strength, 400 steps of many-shot training is not it.

\section{Emergence across OLMo-2 checkpoints}\label{app:olmo}
\begin{figure}[t]\centering
\fig{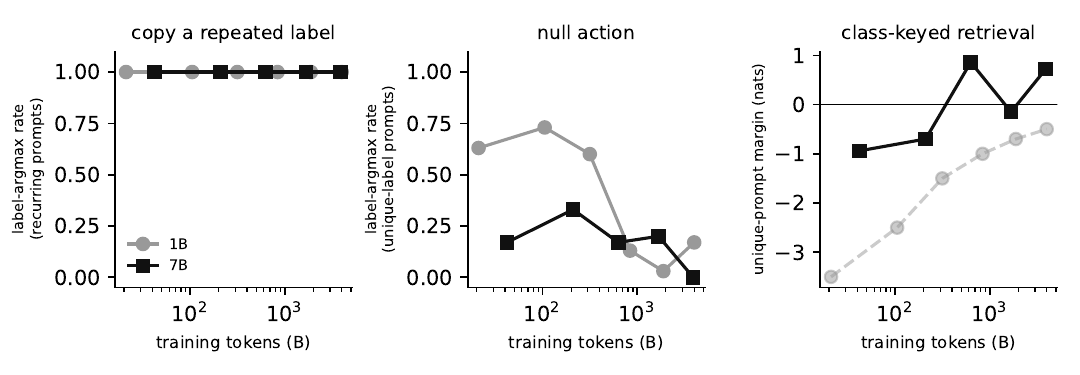}
\caption{The circuit assembles in order during training. Across OLMo-2 stage-1 checkpoints (grey: 1B; black: 7B), copying a repeated label is present from the earliest checkpoint. The null action, measured as withholding a label on unique-label prompts, appears in 1B between 315B and 839B tokens and is already present at 42B in 7B. Class-keyed retrieval, measured as a positive margin on unique-label prompts, arrives last, by 630B tokens in 7B and never in 1B. Thirty prompts per point.}
\label{fig:olmo}
\end{figure}

OLMo-2 releases intermediate checkpoints, which let us ask in what order the parts of the circuit appear. We ran the slotgate battery (two labels used twice against four labels used once, thirty prompts each) on six stage-1 checkpoints of OLMo-2 1B (10k, 50k, 150k, 400k, 900k and 1.9M steps; 21B to 4T tokens) and five of OLMo-2 7B (10k to 929k steps; 42B to 3.9T tokens), and read three measures at each.

Copying a repeated label is present from the first checkpoint in both models. On recurring prompts the label is the argmax at every checkpoint, and knocking out the sixteen duplicate-token heads removes the recurring margin from the first checkpoint on (1B at 10k steps: $+0.11$ to $-1.96$). The null action comes second. In 1B, the share of unique-label prompts that still receive a label is 0.63, 0.73 and 0.60 at the first three checkpoints and 0.13, 0.03 and 0.17 at the last three, so withholding appears between 315B and 839B tokens; in 7B the share is already 0.17 at 42B tokens. Class-keyed retrieval comes last. The unique-prompt margin in 1B is negative at every checkpoint ($-3.5$ to $-0.5$), and the model never does the task. In 7B it turns positive by 630B tokens ($+0.86$, accuracy 0.67; $+0.72$ at 3.9T). The order of copying, then the null action, then the readers is the one the dependencies require.

Two observations are specific to immature checkpoints. In trained 1B and in early 7B, knocking out the duplicate-token heads raises label emission on unique prompts (1B: 0.17 to 0.73), the opposite sign to Qwen; the trained 7B has Qwen's sign. We also ran Pythia-1.4B across six checkpoints, but it is at chance on the task at every one and tells us nothing about the order.

\section{Example prompts}\label{app:prompts}
One verbatim prompt per battery as the Qwen tokenizer sees it. The query sentence is held out of the demonstrations, and the model is scored on the next token after the final arrow.

\paragraph{Twelve demonstrations, invented words (concept12)} 149 tokens; target \texttt{quil}, alternative \texttt{zorp}.
\begin{quote}\small\ttfamily
kludom trolom blidom -> quil\\
tratak grodak gridak -> zorp\\
prunom trogom plutom -> quil\\
drobak klirak vrirak -> zorp\\
blitom frelom klumom -> quil\\
sturak grupak grodak -> zorp\\
klemak tramak blirak -> zorp\\
vromak snibak stenak -> zorp\\
dropom snukom plapom -> quil\\
plisom vrebom drapom -> quil\\
frelak prirak gripak -> zorp\\
pribom vredom vremom -> quil\\
grinom blorom skedom ->
\end{quote}
\paragraph{Four demonstrations, labels repeat (slotgate, recurring member)} 45 tokens; target \texttt{foo}, alternative \texttt{bar}.
\begin{quote}\small\ttfamily
The omelet was folded over melted cheese -> foo\\
a wonderful, heartfelt film -> bar\\
an absolute joy to read -> bar\\
The pancakes were stacked with syrup -> foo\\
The steak rested before it was sliced ->
\end{quote}
\paragraph{Four demonstrations, every label once (slotgate, unique member; token-identical to the recurring member except the two late labels)} 45 tokens; target \texttt{foo}, alternative \texttt{bar}.
\begin{quote}\small\ttfamily
The omelet was folded over melted cheese -> foo\\
a wonderful, heartfelt film -> bar\\
an absolute joy to read -> vex\\
The pancakes were stacked with syrup -> baz\\
The steak rested before it was sliced ->
\end{quote}
\paragraph{Three classes, four demonstrations each (manyclass)} 121 tokens; target \texttt{foo}, alternative \texttt{vex}.
\begin{quote}\small\ttfamily
cold food and a long wait -> vex\\
a wonderful, heartfelt film -> lux\\
Which road leads to the harbor -> foo\\
charming and genuinely funny -> lux\\
What time does the pharmacy close -> foo\\
Should we take the highway or the coast road -> foo\\
buggy, slow and badly designed -> vex\\
wooden acting and a lazy script -> vex\\
cozy room with a great view -> lux\\
the worst concert I have seen -> vex\\
a warm and moving story -> lux\\
Where did you leave the car keys -> foo\\
How long does the bread need to rise ->
\end{quote}

\section{Trained model: the gate condition}\label{app:toy}
We used three training distributions for the 3.4M-parameter model of Section~\ref{sec:build}, all with arbitrary labels redrawn for every prompt. In \emph{mix}, half the prompts are unique-label prompts with a consistent query label and half are recurring-label prompts. \emph{Recurring-only} has only recurring-label prompts. In \emph{gate}, half the prompts are recurring-label, and the other half are unique-label prompts whose query label is drawn fresh from the labels not used in the demonstrations, so that retrieval is never rewarded on a unique-label prompt. Runs lasted 40k steps (mix, recurring-only) or 80k steps (gate; one 40k gate seed that learned is also shown), with three to five seeds each. One gate seed did not learn the task and is excluded. Margins are over 160 held-out prompts per kind.

\begin{table}[h]\centering\small
\begin{tabular}{llcccccc}
\toprule
training & seed & unique & recurring & direction & recurring $\to$ unique & unique $\to$ recurring \\
\midrule
mix & 0 & 8.8 & 8.8 & 0.60 & 8.9 (8.8) & 8.7 (8.7) \\
recurring-only & 0 & 10.3 & 10.4 & 0.54 & 10.4 (10.4) & 10.2 (10.3) \\
gate & 0 & 1.8 & 9.1 & 0.93 & \textbf{0.7} (7.0) & 1.8 (1.5) \\
gate & 1 & 2.3 & 9.0 & 1.00 & \textbf{1.1} (5.8) & \textbf{8.0} (1.8) \\
gate & 4 & 2.0 & 8.7 & 0.99 & 8.1 (7.1) & \textbf{4.0} (0.9) \\
gate & 2$^{\dagger}$ & 2.0 & 8.7 & 0.81 & 7.3 (8.2) & 2.9 (1.9) \\
\bottomrule
\end{tabular}
\caption{The gate forms only when retrieval on unique-label prompts is penalised. ``Unique'' and ``recurring'' are intact margins in nats (correct label against the best other demonstrated label) on held-out prompts of each kind. ``Direction'' is the held-out sign accuracy of the mean-difference direction between recurring- and unique-label prompts at the query arrow after layer 2 (chance 0.5). The last two columns steer along that direction by one natural gap at the arrow in layer 2 and report the resulting margin, with a random direction of the same norm in parentheses: recurring prompts pushed toward ``unique'', and unique prompts pushed toward ``recurring''. Bold: the steer moves the margin more than 3 nats further than its random control. Mix and recurring-only: one seed shown; the other seeds have direction accuracy within the same 0.45--0.62 range and are reported in the released logs. $^{\dagger}$40k steps.}
\label{tab:toygate}
\end{table}

In the gate runs the same slot and reader parts are present as in the other two distributions (a slot swap at the writer layers moves the answer by 9 nats, and one or two last-layer readers carry the class-flip attribution), so the gate is added to an otherwise unchanged mechanism and does not replace any of it. Steering is partial and depends on the seed. In two seeds the direction closes the gate (recurring prompts lose 8 nats, against 2--3 for a random push), in two it opens it (unique prompts gain 2--6 nats, against none), and no seed can be steered fully in both directions at this dose. The direction is a real and necessary part of the signal that opens the read, but not all of it, which matches what we found for $g$ in the pretrained models (Section~\ref{sec:arrows}). An earlier draft of this test used a fixed push of $0.5\sqrt{d}$, which is several natural gaps in a model this small. At that dose a random direction collapses the margin too, so the dose has to be set from the measured gap.

\section{Token observatory: one prompt, every token, every layer}\label{app:trace}
\begin{figure}[t]\centering
\includegraphics[width=\textwidth]{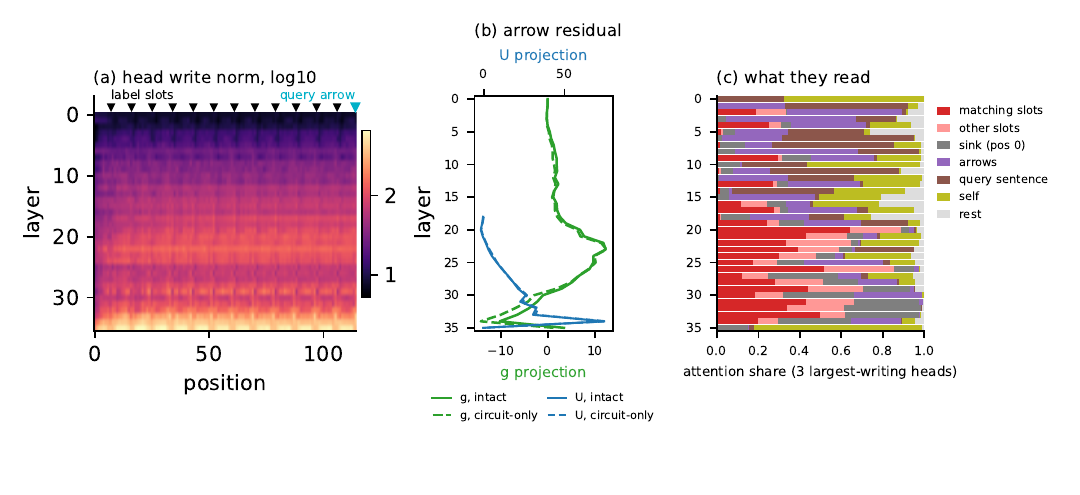}
\caption{Token observatory for one 12-shot prompt (symbol labels, answer \texttt{foo}) in Qwen3-4B. (a)~Summed head write norm at every layer and position (log scale); white lines mark the label slots, and cyan marks the query arrow. (b)~At the query arrow, the residual's projection onto the gate direction $g$ (green) and onto the repeat-flag direction $U_l$ (blue) by layer, intact (solid) and with every unnamed head mean-ablated at the arrow (dashed). (c)~Attention of the three heads that write most at the arrow in each layer, split by what they read.}
\label{fig:trace}
\end{figure}

The knockout numbers above compress a forward pass into one scalar. So that a single token can be followed instead, the released runner has a trace mode. For a given prompt it writes the residual stream after every layer at every position, every head's attention pattern, the norm of every head's write at every position, and the residual's projection onto the gate direction $g$ and the repeat-flag direction $U_l$, to a shard (29\,MB per prompt in half precision; four prompts, intact and circuit-only, are shipped with the viewer). Figure~\ref{fig:trace} reads one such shard.

Panel (a) shows where the heads write. At every layer the label slots and the arrows receive more head output than the sentence tokens, and from layer 20 on the query arrow receives the most. Panel (b) follows the arrow's residual stream. The $g$ projection rises from $+2.5$ at layer 18 to $+12.3$ at layer 23, exactly across the pooling layers, and falls after the readers have used it, while the repeat-flag projection grows from layer 22 on as the readers write label evidence. Panel (c) classifies, layer by layer, what the three heads that write most at the arrow attend to. In the early layers they attend to the query sentence (6:20, 7:31), in the teens to the demonstration arrows (8:23, 19:26), from layer 20 to 29 to the matching label slots (these are the readers), and in layers 29--34 to the sink and the matching slots (brakes and copiers). The circuit-only trace is the same shard with every unnamed head mean-ablated at the arrow. Its residual departs from the intact one by 1 unit at layer 0 and by 134 at layer 34 (17\% of the norm), and its $g$ and $U$ curves are the dashed lines in (b).

\section{Completeness screens}\label{app:complete}
\paragraph{Layers 20--27 in Qwen.} Section~\ref{sec:accounting} traced what the named set lacks to the unnamed heads of layers 20--27 at the query. To name them, we mean-ablated each of the 256 heads in those layers alone at the query (reference: the head's mean output over the battery), on 144 prompts per head, and measured the paired change in margin. The screen tests the named set as well as searching beyond it. The four largest single-head effects all belong to heads already in the named set: the two query heads (23:26 $-0.11$, 22:10 $-0.11$) and the readers 22:13 ($-0.10$) and 24:22 ($-0.09$). Of the 33 named heads in the band, 11 cost more than 0.02, against 17 of the 223 unnamed ones. Of the unnamed heads, 160 change the margin by less than 0.01 in either direction.

Eight unnamed heads have an effect below $-0.03$ with a confidence interval excluding zero. Four of them put 0.48--0.61 of their attention on the recurring-label slots and are readers that the top-20 ranking missed (24:23 $-0.06$, 22:22 $-0.05$, 22:23 $-0.04$, 24:13 $-0.04$). The other four attend almost nowhere on the slots (0.01--0.10), and their attention patterns at the query show what they are. 20:5 and 23:25 put 0.79 and 0.59 of their attention on the query sentence's own tokens, so they are query heads of the kind already named (22:10, 23:26), writing the query's class code into the arrow. 27:12 and 20:17 put 0.55 and 0.34 on the demonstration arrows (and 0.29 / 0.10 on the sink), so they read the prompt-level state that the pooling stage writes at the arrows. Knocking out the four together on a fresh draw costs $-0.25$ / $-0.12$ nats (rare / symbol), against $+0.03$ / $+0.03$ for layer-matched random quadruples. All eight screened heads are instances of parts the account already has, and none is a new stage. A further seven unnamed heads have effects of $-0.015$ to $-0.03$ with intervals excluding zero.

The band's joint cost is larger than the sum of its parts' signed effects (layers 22--23: $-0.36$ / $-0.19$ nats jointly, against a signed sum of single-head effects of $-0.01$ / $-0.08$; the negative effects alone sum to $-0.32$). Heads whose removal helps on its own ($+0.28$ in that band) stop helping once the others are gone. This is what a distributed population looks like, and it has the same shape as the result in Section~\ref{sec:arrows} that amplified readers cannot replace the rest.

Adding the screened heads to the named set (79 heads) and rerunning the band test with the reference matched on label pair (Appendix~\ref{app:reference}) shrinks the gap. Removing every remaining unnamed head of layers 20--27 at the query now costs $-0.10$ / $-0.07$ nats (rare / symbol; before the screen, $-0.67$ / $-0.34$), and the unnamed heads of 28--35 help when removed ($+0.17$ / $+0.09$), so outside our set the late heads are, on average, brakes. Keeping only the 79-head set preserves about 60\% of the margin on both label kinds: $-0.47$ / $-0.28$ under resample ablation (intact $+1.14$ / $+0.55$; random set $-0.98$ / $-0.65$, no heads $-0.95$ / $-0.69$), and $-0.26$ / $-0.22$ with the mean reference. The rest is split between the unnamed heads of layers 0--19 at the query ($-0.29$ / $-0.16$ in the band test), the encoder that computes the query sentence's class code at the arrow, which this account treats as the circuit's input and not as one of its parts, and a thin tail in layers 20--27. The natural-label screen of Section~\ref{sec:transfer} names that tail together with the late brakes (110 heads), and the final 189-head set is above intact on this battery too ($+1.26$ / $+0.89$ under resample, $+1.62$ / $+0.93$ with the mean reference; random $-0.98$ / $-0.64$). With it, layers 20--27 and 28--35 help when their leftover heads go ($+0.99$ / $+0.40$, $+0.65$ / $+0.35$), and layers 0--19 cost $-0.17$ / $-0.07$. An earlier version of this test used the battery mean over both label pairs and a resample partner one row on, which in this battery is the same prompt under the other pair. It read $+0.49$ on symbol labels for the 79-head set, and that number came from the partner. The attention patterns of all eight heads at the query are in the released data.

\paragraph{Layers 0--19 in Qwen.} The same screen over the 640 heads of layers 0--19 at the query (five containers, 144 prompts per head) finds no part. One head crosses $-0.03$ (8:23, $-0.037$), three more lie between $-0.02$ and $-0.03$ (19:24, 6:20, 6:15), and the remaining 636 are within $\pm0.02$ of zero; only two help when removed. The band knockouts are small and shrink with depth: layers 0--3 $+0.00$ / $-0.01$, 4--7 $-0.09$ / $-0.05$, 8--11 $-0.05$ / $-0.02$, 12--15 $-0.02$ / $-0.02$, 16--19 $-0.11$ / $-0.06$ (rare / symbol; the intervals for 0--3, 12--15 and the symbol value of 16--19 include zero). The 16--19 band holds the one early head that looks like a reader (19:11, with 0.89 of its attention on the recurring slots, $-0.015$) and the second-largest early effect (19:24, $-0.029$, which attends to neither the slots nor the sink). Knocking out the four largest early heads together (8:23, 6:20, 6:15, 7:31) on a fresh draw costs $-0.06$ / $-0.02$, against $0.00$ for random quadruples.

The attention patterns of these four heads show how little they do. 8:23 puts 0.97 of its attention on the demonstration arrows (an early reader of prompt-level state, the kind of input the pooling stage uses), 6:20 and 7:31 put 0.97 and 0.68 on the query sentence (early copies of the query's words into the arrow, upstream of the query heads), and 6:15 is mixed. Before reading them we had predicted that the early heads that matter for rare labels would attend to the label slots and carry a label-type feature. They do not, so we dropped that hypothesis, and the rare-label remainder stays unexplained at the level of parts. This is what the gain bound predicts for heads whose output-value map has little component along a late answer direction: the early layers contribute to the margin as a substrate, a few hundredths of a nat per head, and not as parts. Adding those four to the named set together with the eight screened heads (79 heads) does not close the early band ($-0.35$ / $-0.18$ when every other layer 0--19 head is removed at the query, against $-0.38$ / $-0.16$ before), because the early contribution is the sum of six hundred effects of a few thousandths each and not a part one could name. With the 91-head set, the circuit alone keeps 65\% of the rare-label margin ($-0.40$ from $+1.13$; random set $-0.75$, no heads $-0.72$), and on symbol labels it is indistinguishable from intact under battery means ($-0.19$, interval includes zero) and above intact under resample ($+0.46$).

\paragraph{SmolLM3 and Llama.} We ran the same protocol in SmolLM3-3B and Llama-3.2-3B with each model's own named sets (its readers, brakes and copiers from the direct-contribution census, its own duplicate-token heads and its late MLPs; 95 and 94 heads). The first pass was run before the reference was matched on label pair, so its values are only indicative, but it gave the same picture as in Qwen. The named parts carried most of the symbol-label behaviour under resample ablation (SmolLM3 $+0.37$ against $+0.77$ intact, random set $-0.73$; Llama $+0.05$ against $+0.41$, random $-0.01$) but not the rare-label margin ($-0.72$ and $-1.00$), and the band complement located the gap in layers 18--23 and 10--17 in SmolLM3 and layers 8--13 and 14--19 in Llama, with everything later closed in both. Screening those bands head by head (352 and 288 heads) named 12 and 17 unnamed heads with effects below $-0.03$ and intervals excluding zero.

Their attention patterns at the query sort them into parts the account already has. In SmolLM3 there are four readers (18:7, 19:0, 19:1, 20:11; 0.29--0.44 of their attention on the label slots), seven query heads (17:9, 20:6, 20:7, 20:15, 22:5, 22:10, 22:11; 0.34--0.74 on the query sentence) and one sink head (22:4). In Llama there are seven readers (12:2, 12:3, 13:17, 13:23, 14:14, 15:18, 15:19) and eight query heads (11:1, 11:9, 11:19, 12:5, 13:5, 13:6, 14:6, 15:11), so Llama's writer, query and first-reader stage sits at layers 11--15 of 28, earlier than its census readers, and the top-20 reader cut had under-counted it. Knocking out the non-reader additions jointly on a fresh draw costs $-0.68$ / $-0.48$ (SmolLM3) and $-0.50$ / $-0.15$ (Llama), against $-0.02$ / $-0.12$ and $0.00$ / $+0.01$ for random sets.

With the additions (107 and 111 heads) and the reference matched on label set (Appendix~\ref{app:reference}), the circuit alone reproduces the behaviour in both models. Llama is $+0.84$ / $+0.25$ over intact with the mean reference and $+0.68$ / $+0.09$ under resample (random set $-0.77$ / $-0.30$, no heads $-0.80$ / $-0.43$), with every layer band closed ($+0.01$ to $+0.29$ when all other heads of a band are removed). SmolLM3 is $+1.77$ / $+1.19$ with the mean reference and $+0.59$ / $+0.62$ under resample (random $-1.07$ / $-0.74$). Its layers 30--35 are closed ($+0.02$ / $0.00$), the unnamed heads of 18--29 act as strong brakes ($+2.20$ / $+1.45$ when removed), and its layers 0--17 keep the same diffuse substrate cost as Qwen's early layers ($-0.15$ / $-0.13$). With the heads that the natural-label screen adds (Section~\ref{sec:transfer}; 122 and 163 heads) the margins rise again (SmolLM3 $+1.01$ / $+0.87$ under resample, Llama $+0.75$ / $+0.03$; random $-1.11$ / $-0.71$ and $-0.71$ / $-0.26$), and every band except the early encoder band helps or is neutral when its leftover heads go. In the sense of Section~\ref{sec:faithful}, the circuit is complete in all three families.

\section{Choosing the ablation reference}\label{app:reference}
\paragraph{Why resampling is harsher on word labels.} Under resampling, the circuit-only margin is $0.5$--$0.9$ nats lower than under the mean reference on natural labels, but not on symbols. The resample partner is another prompt of the same kind, whose answer is the other label half the time, so resampling only costs something if the unnamed heads carry evidence about the answer. On word labels they do (the screen's unnamed effects sum to $-10$ nats, against $-3$ for the named heads), and on symbols they do not. A word like \emph{animal} can be moved to the logits by many weak heads, while an arbitrary token can be moved only by the few readers that will copy anything. On word labels, then, the reader stage is a population whose active membership depends on what has to be copied, and completeness is a statement about that population.

\paragraph{A control we first got wrong.} The batteries store each prompt several times in a row, once per label set. Our first resample control took its reference from the next prompt, and our mean reference averaged over all the label pairs in the battery. On the natural-label battery, the next prompt after a natural-label prompt is the same sentences with the labels exchanged, so the control injected the opposite mapping. On the symbol battery of Section~\ref{sec:accounting}, the next prompt is the same sentences under the other invented pair, which has the same mapping and can only help. The bias therefore has no fixed sign. When we matched the reference on label set (the mean over same-kind prompts, and the next prompt of the same kind as the partner), one result reversed (SmolLM3 carrying $R$ with its original set), Qwen's symbol-battery circuit-only value fell from $+0.49$ to $-0.28$, and SmolLM3's and Llama's rose. Every number in this paper that compares a circuit-only run with intact now uses the matched reference, and Appendix~\ref{app:complete} keeps the one superseded value with its explanation.

\section{Further connection tests}\label{app:arrows}
\paragraph{Where the rest of $g$ goes.} The readers' queries carry a quarter to a third of $g$'s effect (about half in the run used for this sweep), so we swept for the remainder in Qwen. We removed $g$ from every head's queries in a band of layers, or from the MLP inputs in a band, with the gate direction added at layer 23 and the duplicate heads knocked out. Removing it from all queries in L23--35 takes back as much as removing it from the twenty readers alone ($+0.34 \to +0.18$ in both cases), and the band-by-band values (L23--26 $+0.35$, L27--30 $+0.13$, L31--35 $+0.33$) show that the query-side route runs through the middle band, where the readers sit. Removing $g$ from MLP inputs changes nothing specific ($+0.25$ to $+0.51$ across bands, against $+0.32$ for a random direction). The other half of $g$'s effect is therefore not a second named route through the MLPs. Either it is carried by the same reader queries in a way that direction removal does not reach, or it is carried by heads outside the twenty that each read $g$ weakly, and this design cannot tell the two apart.

\paragraph{A cleaner sink test.} Zeroing position zero's value only as it is read by other positions, while leaving its own residual stream untouched, costs $0.34$ nats on unique-label prompts and $0.05$ on recurring ones in the reader layers ($0.50$ and $0.05$ at every layer), against $0.33$ and $0.03$ for the ordinary zeroing. The sink's effect on the readers is therefore in its value as read by other positions, and what the all-layer zeroing added ($0.76$ against $0.50$) came from position zero's own residual stream. Swapping position zero's value for another prompt's costs $1.50$ nats, so the value is not inert. It is the content the readers fall back on.

\section{Components or populations}\label{app:populations}
Of the links in Table~\ref{tab:replication}, the two query heads, the sixteen duplicate-token heads, the three value groups, SmolLM3's three order carriers and the four-demonstration label-prior neurons are small named sets. The writers, the 89 slot-reading readers, the roughly thirty heads released by the gate and the 24 brakes are populations in which no single unit is necessary. The population links are forced by gain. A reader's class bias is 0.2--0.4 nats on a 0.5-nat per-slot spread, and no head's output writes more than 0.2 nats of answer logit, so a 1.3-nat margin is a sum over at least seven heads (a flip of that size over at least four) and in practice over tens. The units that fill a role change from model to model while the function stays the same (Section~\ref{sec:replication}). The parts of the account that the architecture forces, the null action and the within-regime read law, follow from the softmax and the causal mask and appear in every model we tested.
\end{document}